\documentclass[preprint]{IEEEtran}
\IEEEoverridecommandlockouts
\usepackage[utf8]{inputenc}
\usepackage[T1]{fontenc}

\usepackage{cite}
\usepackage{amsmath,amssymb,amsfonts}
\usepackage{algorithm, algorithmic}
\usepackage{graphicx}
\usepackage{textcomp}
\usepackage{xcolor}
\usepackage{hyperref}
\usepackage{tcolorbox}
\usepackage{threeparttable}
\usepackage{booktabs}
\usepackage{makecell}
\usepackage{todonotes}
\usepackage[normalem]{ulem} 
\usepackage{multirow}
\usepackage{enumitem}
\usepackage{mathtools}
\usepackage{amsthm}
\usepackage[capitalize, noabbrev]{cleveref}
 
\def\BibTeX{{\rm B\kern-.05em{\sc i\kern-.025em b}\kern-.08em
    T\kern-.1667em\lower.7ex\hbox{E}\kern-.125emX}}

\theoremstyle{plain}

\theoremstyle{definition}

\theoremstyle{remark}

\newcommand{\nameShort}{K-FADE}
\newcommand{\vect}[1]{\mathrm{vec}\left(#1\right)}

\newcommand{\pgrad}[1]{\mathcal{D}{#1}}

\newtcolorbox{reviewquote}{
    colback=gray!10,
    colframe=gray!30,
    boxrule=0.5pt,
    left=4pt,
    right=4pt,
    top=4pt,
    bottom=4pt,
    arc=2pt,
    boxsep=0pt,
    fontupper=\itshape
}

\title{Gauss-Newton Unlearning for the LLM Era}

\makeatletter 
\newcommand{\linebreakand}{%
  \end{@IEEEauthorhalign}
  \hfill\mbox{}\par
  \mbox{}\hfill\begin{@IEEEauthorhalign}
}
\makeatother 

\author{\IEEEauthorblockN{Lev McKinney\IEEEauthorrefmark{2}\IEEEauthorrefmark{4}, Anvith Thudi\IEEEauthorrefmark{2}\IEEEauthorrefmark{4}, Juhan Bae\IEEEauthorrefmark{2}, Tara Rezaei\IEEEauthorrefmark{3},\\Nicolas Papernot\IEEEauthorrefmark{2}\IEEEauthorrefmark{4}, Sheila A. McIlraith\IEEEauthorrefmark{2}\IEEEauthorrefmark{4}, and Roger Grosse\IEEEauthorrefmark{2}}
\IEEEauthorblockA{\IEEEauthorrefmark{2}University of Toronto, Ontario, Canada}
\IEEEauthorblockA{\IEEEauthorrefmark{3}MIT, Massachusetts, USA}
\IEEEauthorblockA{\IEEEauthorrefmark{4}Vector Institute for Artificial Intelligence, Canada}
}
\IEEEpubid{
\parbox{\columnwidth}{\vspace{2em}\centering\normalfont\scriptsize This work has been accepted for publication at the IEEE Conference on Secure and Trustworthy Machine Learning (SaTML). The final version will be available on IEEE Xplore.}
\hspace{\columnsep}\makebox[\columnwidth]{}
}

\begin{document}

\maketitle

\newcommand{\showme}[1]{{\color{magenta} #1}}
\newcommand{\notes}[1]{{\color{cyan} #1}}
\newcommand{\revisit}[1]{{\color{blue} #1}}   
\newcommand{\added}[1]{{\color{red} #1}}
\newcommand{\alt}[1]{{\color{brown} $\backslash$#1}}
\newcommand{\remove}[1]{{\color{green} #1}}
\newcommand{\removeifneeded}[1]{{\color{cyan} #1}}
\newcommand{\removehide}[1]{}
\newcommand{\todosm}[1]{\textcolor{blue}{({\bf SM TODO:} #1)}}

 \newcommand{\nocommentsm}[1]{}
 \newcommand{\addCRC}[1]{}


\newif\ifcomments
\commentstrue
\ifcomments

    \newcommand{\commentsm}[1]{\textcolor{cyan}{({\bf SM:} #1)}}
    \newcommand{\commentsmhide}[1]{}
    
    \newcommand{\commentlm}[1]{\textcolor{orange}{({\bf LM:} #1)}}
    \newcommand{\commentlmhide}[1]{}
    
    \newcommand{\commentrg}[1]{\textcolor{teal}{({\bf RG:} #1)}}
    \newcommand{\commentrghide}[1]{}
    
    \newcommand{\commentjb}[1]{\textcolor{magenta}{({\bf JB:} #1)}}
    \newcommand{\commentjbhide}[1]{}

    \newcommand{\commentat}[1]{\textcolor{blue}{({\bf AT:} #1)}}
    \newcommand{\commentathide}[1]{}

    \newcommand{\commenttr}[1]{\textcolor{red}{({\bf TR:} #1)}}
    \newcommand{\commenttrhide}[1]{}

    \newcommand{\commentnj}[1]{\textcolor{brown}{({\bf NickJ:} #1)}}
    \newcommand{\commentnjhide}[1]{}

\else

    \newcommand{\commentsm}[1]{}
    \newcommand{\commentsmhide}[1]{}
    
    \newcommand{\commentlm}[1]{}
    \newcommand{\commentlmhide}[1]{}
    
    \newcommand{\commentrg}[1]{}
    \newcommand{\commentrghide}[1]{}
    
    \newcommand{\commentjb}[1]{}
    \newcommand{\commentjbhide}[1]{}

    \newcommand{\commentat}[1]{}
    \newcommand{\commentathide}[1]{}
    
    \newcommand{\commenttr}[1]{}
    \newcommand{\commenttrhide}[1]{}

    \newcommand{\commentnj}[1]{}
    \newcommand{\commentnjhide}[1]{}
\fi

\begin{abstract} 
Standard large language model training can create models that produce outputs their trainer deems unacceptable in deployment. The probability of these outputs can be reduced using methods such as LLM unlearning. However, unlearning a set of data (called the forget set) can degrade model performance on other distributions where the trainer wants to retain the model's behavior. To improve this trade-off, we demonstrate that using the forget set to compute only a few uphill Gauss-Newton steps provides a conceptually simple, state-of-the-art unlearning approach for LLMs. While Gauss-Newton steps adapt Newton's method to non-linear models, it is non-trivial to efficiently and accurately compute such steps for LLMs. Hence, our approach crucially relies on parametric Hessian approximations such as Kronecker-Factored Approximate Curvature (K-FAC). We call this combined approach K-FADE (\textbf{K-FAC} for \textbf{D}istribution \textbf{E}rasure). Our evaluation on the WMDP and ToFU benchmarks demonstrates that K-FADE suppresses outputs from the forget set and approximates, \textit{in output space}, the results of retraining without the forget set. Critically, our method does this while altering the outputs on the retain set less than previous methods. This is because K-FADE transforms a constraint on the model's outputs across the entire retain set into a constraint on the model's weights, allowing the algorithm to minimally change the model's behavior on the retain set at each step. Moreover, the unlearning updates computed by K-FADE can be reapplied later if the model undergoes further training, allowing unlearning to be cheaply maintained.
\end{abstract}




\section{Introduction}
Large Language Models (LLMs) draw many of their capabilities from training on enormous swaths of the internet. The size of these training sets and their unstructured nature makes curating them difficult. Thus models often learn to produce outputs outside the intended use-case of the model developer. Models can memorize sensitive information~\cite{huang-etal-2022-large, carlini2021extracting}, like emails and phone numbers, or produce information that may enable harmful activities such as the construction of chemical, biological, radiological, and nuclear (CBRN) weapons~\cite{li_wmdp_2024}. It is also possible for adversaries to poison common LLM training datasets~\cite{carlini2024poisoning}. Unlearning for LLMs aims to address this by modify a model to prevent it from readily producing these undesirable outputs and reducing the effect of the training data that caused it \cite{liu_rethinking_2024}.

We operationalize this LLM unlearning problem by casting it as the solution to a multi-objective optimization. The first part of the objective is to reduce the probability of outputs from a forget distribution, represented by a ``forget set". We call this desideratum \emph{output suppression}~\cite{cooper2024machine}. This objective must be achieved while being constrained to minimally alter the model's outputs on the distribution we want to perform well on; we call this desideratum \emph{specificity}.\footnote{This is analogous to specificity in classification, where here data outside the forget set is our negative class, and we want this unchanged.} While these are our primary objectives, this operationalization does not cover all of the objectives a model developer may want from an unlearning method. Thus, we consider whether our unlearning method can mimic the outputs of a model trained with the forget distribution excluded (approximate retraining) and whether it can resist finetuning attacks that attempt to recover the unlearned behavior (\emph{robustness}) \cite{cooper2024machine, liu_rethinking_2024}.

Just optimizing this initial operationalization is difficult. Many methods attempt to directly tackle suppressing outputs under specificity constraints with variants of gradient difference algorithms~\cite{maini_tofu_2024, yao_large_2024, yao_machine_2024, elm2024, jia_soul_2024, zhang2024npo, fan2024simplicity}. These iterative algorithms require careful tuning to ensure their stability and necessarily use relatively small samples to estimate how each step of their algorithm changes the model's behavior on the retain distribution. This methodology has limited how well the LLM unlearning field has been able to achieve output suppression with high specificity.

We first observe that we can simplify the optimization process by embedding the specificity constraint into the optimizer itself. We do this by taking a quadratic approximation to our specificity constraint (KL divergence), a linear approximation to our suppression objective, and then solving the subsequent linear system. This method is generally called the Gauss-Newton update, and in our setting allows output suppression and specificity to be simultaneously optimized by following a single update direction. Overall, this provides a much more stable optimization algorithm.

However, the Gauss-Newton update requires computing inverse-Hessian-vector products; we hence show how to efficiently and accurately estimate these for the purposes of unlearning. Past work on unlearning with second-order methods in LLMs used diagonal approximations to the Hessian \cite{jia_soul_2024}, but this is often an inaccurate approximation, since it ignores any interactions between the parameters.
We instead build off of work on parametric estimation of Hessians, e.g., Kronecker-Factored Approximate Curvature (K-FAC)~\cite{martensOptimizingNeuralNetworks2015} and Eigenvalue-corrected K-FAC (EK-FAC)~\cite{george2018fast}, and work demonstrating these methods' efficacy in LLMs~\cite{grosse2023studyinglargelanguagemodel}. With these advances, and a careful study into how to scale the Gauss-Newton step size, we create an approach that can efficiently and effectively achieve both output suppression and high specificity. We call our method \textbf{K-FA}C for \textbf{D}istribution \textbf{E}rasure (K-FADE). We found K-FADE's runtime can be much faster than re-training, and can be on par with the runtime of past first-order methods while still outperforming them.

Our experiments across benchmarks show that K-FADE can improve the Pareto frontier of LLM unlearning along the output suppression and specificity axes, and that it can mimic the effect of removing fine-tuning data. We first evaluate on the Weapons of Mass Destruction Proxy (WMDP) benchmark~\cite{li_wmdp_2024}, which measures a method's ability to suppress proxies for ``hazardous'' content while maintaining broad knowledge and fluency as measured by the MMLU~\cite{hendrycks2021measuring} and MT-Bench~\cite{zheng2023judging} benchmarks. After unlearning on WMDP, we show that \nameShort{} achieves strong results in output suppression, while maintaining performance on the other benchmarks.
Second, we evaluate it on the Test of Fictitious Unlearning (ToFU)~\cite{maini_tofu_2024}, which benchmarks an unlearning method's ability to remove sensitive information about individuals while preserving nonsensitive information and causing the model to mimic the output distribution of a model trained without this sensitive data.
We find that a single Gauss-Newton step delivers a Pareto improvement in Forget Quality, i.e., approximate retraining metrics, and model utility, representing a new state-of-the-art on the ToFU benchmark.

We also perform a more fine-grained analysis of specificity on instruction following data. LLMs are typically deployed as chatbots where they are used in diverse settings. Hence, maintaining performance on this distribution is paramount to deployability. This motivates us to study the KL-divergence of the outputs between the unlearned model and the original model across instruction-following data~\cite{alpaca}. We find that the KL divergence caused by K-FADE is generally smaller on this instruction following set even though it was never explicitly used to fit the Hessian defining our specificity constraint. 

In exploring our method's improvement across unlearning benchmarks, we also investigated the efficiency of maintaining unlearning over the life cycle of a model. A common and effective attack against LLM unlearning is simply fine-tuning the unlearned model, which has been shown to remove the effect of unlearning~\cite{deeb2024unlearningmethodsremoveinformation, qi2024evaluating, rosati2024representation}. This is a threat even for closed-source models when a service also provides a fine-tuning API for the LLM, or for developers who want to continue to train their unlearned models without losing the unlearning effect. While fine-tuning on various datasets still degrades the strength of unlearning from K-FADE, we empirically found that the previously computed K-FADE unlearning update often transferred to the fine-tuned model and consistently performed better—at unlearning—than transferring the unlearning update from other methods. Hence, a fine-tuning API can re-apply the previously computed K-FADE update to maintain unlearning cheaply, or a model trainer who updates their own model can simply re-apply the old unlearning update to maintain the unlearning objective.

Ultimately, in this paper we provide an LLM unlearning method which improves the Pareto frontier on several datasets, with respect to several metrics. However, we do not provide formal guarantees for the quality of unlearning when using this method, which can be necessary in some use cases. Moreover, the metrics we evaluated on may not reflect the notions of model utility and output suppression important to every deployment. Beyond this, the threat model for LLM unlearning is that a model trainer is trying to maintain viable utility while scoping the outputs of the model; this does not cover the traditional machine unlearning threat model of data providers revoking access to their data (where one cannot maintain utility if it opposes unlearning effectiveness).

\begin{figure*}
    \centering
    \includegraphics[width=0.8\linewidth]{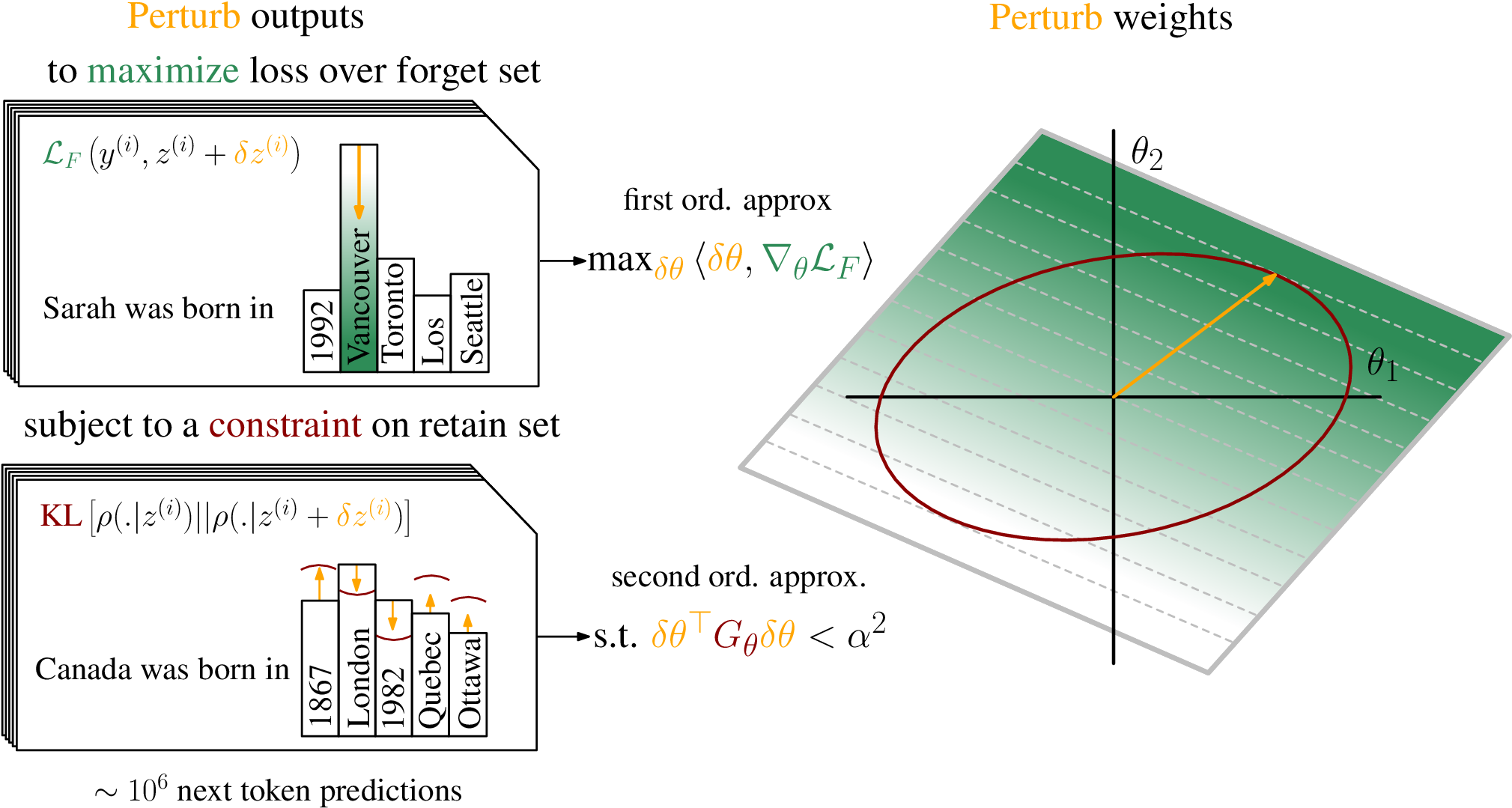}
    \caption{\textbf{How Gauss-Newton ascent achieves output suppression while maintaining performance on a retain set.} Here we show how  Gauss-Newton ascent relaxes a constraint satisfaction problem for unlearning in the output space of a model (represented by the logits $z$) and converts it into a simple (though very high dimensional) linear algebra problem in weight space (represented by $\theta$ variables). Here $\rho(\cdot|z)$ denotes the distribution of tokens (i.e., labels) specified by the model output logits. This allows us to simultaneously satisfy a constraint on the model's output across millions of tokens at every step of the algorithm. See Section~\ref{sec:natgrad} for more details.}
    \label{fig:main_figure}
\end{figure*}

\section{Preliminaries}
\label{sec:prelim}

We now go over our threat model for LLM unlearning, the two objectives for unlearning we consider, and highlight how the Gauss-Newton step is relevant to both of them. To do so we connect observations from several different works, spanning approximate retraining to output suppression. The high-level theme is that to find the best "local" update for unlearning, we can take a linear approximation to the output suppression objective and a second-order approximation to the specificity constraint, and find the minimizer of this local loss function. This is depicted in Figure~\ref{fig:main_figure}.

\subsection{Our LLM Unlearning Threat Models and Notation}

We now state our threat model for LLM unlearning. The model trainer aims to train a language model which takes tokens $x$ as an input and returns a set of unnormalized log probabilities (\textit{logits}) over the next token $z$ as an output: we denote this parametric function as $z = f(x; \theta_D)$ where $\theta_D$ are the parameters coming from training on dataset $D$\footnote{Typically these parameters are not deterministic given a dataset, as they depend on initialization and data ordering. For our analysis below you can assume these variables are also fixed.}. The adversary is an entity that can manipulate parts of the training data, with the goal of changing the trained model's outputs. This exploit is achievable given the scale of current pre-training corpora used for production LLMs~\cite{carlini2024poisoning, souly2025poisoningattacksllmsrequire}. 

If a model trainer discovers such an exploit, we will assume that they have access to a finite \textit{forget set} of samples from the undesired distribution, $D_F$. The model trainer then wishes to reduce the probability of producing outputs in $D_F$, but do so "minimally" as to maintain the other objectives for training (Section~\ref{sec:prob_setting} describes two common objectives for a minimal change). Practically, this threat model captures the common occurrence of LLMs trained on parts of the internet learning to produce outputs not originally intended by the model trainer.

Note this threat model differs somewhat from the traditional machine unlearning setting, where data providers request the model trainer to remove their data~\cite{bourtoule2021machine}. In our threat model the trainer is initiating the "forgetting" of some dataset $D_F$.

Later in the paper we consider an additional unlearning threat model, which is maintaining unlearning of $D_F$ after fine-tuning on (potentially adversarial) data. Here we focus on a fine-tuning API, where the model parameters must be updated to perform well on a requested (potentially adversarial) fine-tuning dataset yet needs to still maintain unlearning of the forget set $D_F$. The primary question will be how efficiently one can re-apply unlearning on the fine-tuned model, i.e., if maintaining unlearning behind a fine-tuning API is viable.

\subsection{Problem Settings}
\label{sec:prob_setting}

In this paper, we primarily focus on using unlearning methods to achieve \textit{output suppression} while minimally changing model behavior outside the forget distribution. In some cases, \emph{we know precisely which training data} caused a generative model to produce particular sensitive outputs (e.g., obscure facts about individuals or memorized text~\cite{carlini2021extracting}). For example, in the ToFU~\cite{maini_tofu_2024} benchmark, models explicitly finetune on facts about fictitious authors which we then want to "forget". In such cases, the forget set is clear, and approximating the output distribution of a model never trained on this data (\textit{approximate retraining}) allows us to maintain privacy and should also achieve our objective of output suppression. Formally, letting $D_R = D \setminus D_F$ be the dataset we want to retain, $\theta_{D_R}$ be the model weights coming from training on $D_R$, the objective is to find some model parameters $\theta$ s.t 

\begin{equation*}
    \theta \approx  \theta_{D_R}\tag{Approximate Retraining}
\end{equation*}

However, not all cases are so clear-cut. ML practitioners often don't know exactly which training inputs led to a particular higher-order concept or capability.\footnote{In fact, we cannot generally determine if a model was trained on a particular input~\cite{thudi2022necessity,thudi2024gradients,cooper2024machine}.} In these cases we can still aim to eliminate the model's capability while minimally affecting its utility~\cite{liu_rethinking_2024, li_wmdp_2024}, regardless of what data originally caused the model to learn it. We do this by reducing the probability of generating the outputs in the forget set $D_F$, while minimally distorting the outputs on relevant use-cases, represented again by a retain set $D_R$ but no longer necessarily the same as $D \setminus D_F$. Formally, we will let $\mathcal L_F(\theta)= \mathbb{E}_{(x, y)\sim D_F}[\ell_F(f(x;\theta), y)]$ be a function such that maximizing it lowers the probability of producing datapoints in $D_F$. Letting $\textrm{KL}(p,q)$ be the KL-divergence between two distributions, $\rho(.|f(x; \theta))$ be the distribution of outputs defined by the model logits $z = f(x; \theta)$, and $\gamma$ be a hyper-parameter, the objective for LLM unlearning in this setting is

\begin{multline*}
    \arg \max_{\theta}  \mathcal L_{F}(\theta) - \gamma \underset{x \sim D_R}{\mathbb{E}} \left[\textrm{KL}(\rho(.|f(x; \theta_D)), \rho(.|f(x; \theta))\right] \tag{Output Suppression}\label{eq:penalty}
\end{multline*}

It is common to use KL divergence regularization to impose the constraint of minimally altering the model outputs on the retain dataset $D_R$ ~\cite{kurmanji2023towards, elm2024, maini_tofu_2024, liu2022continual}.

In the rest of the preliminaries we show how these two objectives are both approximately solved with the same change to the original model weights. This is given by the Gauss-Newton pre-conditioned gradient, also commonly called the "natural gradient". Formally, if we let $G_\theta$ be the Gauss-Newton Hessian for the family of output distribution specified by $f(x;\theta)$ for each $x \in D_R$\footnote{The precise formula is for each $f(x;\theta)$ we can define the Fisher information matrix for the distribution of outputs as $F_{\theta}(x)$. In our setting the Gauss-Newton is then the expectation of these matrices over $x \sim D_{R}$. Section~\ref{sec:hessian_estimation} describes methods to compute this}, and $\eta$ a step-size hyperparameter, then the Gauss-Newton step is

\begin{equation}
    \theta + \eta G_{\theta}^{-1} \nabla_\theta \mathcal{L}_{F}(\theta) \tag{Gauss-Newton Ascent Step}
\end{equation}

We will show that starting from the original model $\theta_D$ and applying the Gauss-Newton step provides a method to approximately solve both of the unlearning objectives we described earlier.

\subsection{Output Suppression and the Gauss-Newton Step}
\label{sec:natgrad}

Gauss-Newton steps can also be seen as increasing the objective on the current sample while minimally changing the models outputs elsewhere \cite{amari1998natural, martensNewInsightsPerspectives2020}. Specifically, instead of directly optimizing~\cref{eq:penalty} with an iterative method, such as stochastic gradient ascent on $\mathcal L_{F}(\theta)$ and descent on the KL term, we can use first-order information about the loss and second-order information about the KL divergence term to accurately estimate the single update in the neighborhood of $\theta$ that achieves both of these objectives simultaneously, depicted in Figure~\ref{fig:main_figure}.

First, rewriting the KL divergence of our model outputs on the retain set with respect to a perturbation $\delta \theta$ of the parameters, and using the second order Taylor expansion\footnote{The first order term is $0$ due to the optimality of $\delta \theta = 0$ for KL divergence.}, we get
\vspace{-0.1em}
\begin{align}
    &\mathbb{E}_{x \sim D_R}[\textrm{KL}(\rho(.|f(x; \theta)), \rho(.|f(x; \theta + \delta\theta))]\\
    &= \frac{1}{2}\delta \theta^\top G_\theta \delta \theta + o(||\delta \theta||^2) \label{eq:kl_approx}
\end{align}

Where $G_\theta$ is the Gauss-Newton Hessian as described earlier in Section~\ref{sec:prob_setting}. Notice that this approximation improves for smaller perturbations $\delta \theta$.

Since this second order approximation (Equation \ref{eq:kl_approx}) to our KL divergence is convex, we can use it and rewrite the penalized optimization in Eq \ref{eq:penalty} as a constrained optimization problem and consider the best update direction

\begin{equation}
    \frac{1}{\epsilon} \underset{\delta \theta \text{ s.t. } \delta \theta^\top G_\theta \delta \theta \leq \epsilon^2}{\textrm{argmax}} \mathcal L_F(\theta + \delta \theta) \label{eq:const_opt}
\end{equation}

If we assume $\mathcal L_F(\theta + \delta \theta)$ is maximized at the boundary $\delta \theta^\top G_\theta \delta \theta = \epsilon^2$, then the solution\footnote{which can be derived using the method of Lagrange multipliers} for the best direction is simply,

\begin{equation}
    \bar{\nabla}_\theta \mathcal{L}_F(\theta) := -\frac{G_\theta^{-1} \nabla_\theta \mathcal{L}(\theta)}{||\nabla\mathcal{L}_\theta||_{G_\theta^{-1}}} \label{eq:natgrad_def}
\end{equation}

where $||v||_M = \sqrt{v^\top M v}$. In general as $\epsilon \rightarrow 0^+$, the best direction will exactly be the above $\bar{\nabla} \mathcal{L}_F(\theta)$, also known as the \emph{natural gradient}~\cite{amari1998natural, martensNewInsightsPerspectives2020}. Furthermore, note $\bar{\nabla} \mathcal{L}_F(\theta)$ is proportional to the Gauss-Newton step and in fact we normalize by $\frac1{||\nabla\mathcal{L}_\theta||_{G_\theta^{-1}}}$ in Sec \ref{sec:methods} to ensure each step with a certain learning rate causes an approximately constant change in KL divergence.


Using the Gauss-Newton Hessian in this way has the advantage of allowing us to consider effects on the KL divergence on the entire retain set with each step. However, $G_\theta$ is often close to singular, leading to numerical instability and poor performance. Thus, we typically introduce a small \textit{damping} term $\lambda$, making the complete update $(G_\theta + \lambda I)^{-1} \nabla_\theta \mathcal{L}(\theta)$. This can also be interpreted as adding an additional l2 distance penalty on how far the weights can travel from the base model to Equation \eqref{eq:penalty} \cite{bae2022if}.

\subsection{Approximate Retraining and the Gauss-Newton Step}
\label{sec:approximate_retraining}

The Gauss-Newton ascent steps we use in this paper also serve as a formal approximate retraining (often also called approximate unlearning) algorithm in linear models as shown in~\cite{guo2019certified}; while the theory does not hold for neural networks, by linearizing a neural network locally we may still apply a similar technique, though without theoretical guarantees. The following derivation serves to provide additional motivation, but is not necessary for understanding our methodology or empirical results.

First let's recall that approximate retraining involves splitting a dataset $D$ into a forget set $D_F$ and a retain set $D_{R} = D \setminus D_F$. With a training process $\theta_D = \mathcal{T}(D)$, unlearning then aims to approximate retraining the model only on the retain set, $\theta_{D_R} = \mathcal{T}(D_R)$, without incurring its associated costs.

For linear models $z=\theta^\top x$ that minimize a strictly convex loss $\ell(z, y)$, effective approximate unlearning becomes relatively straightforward~\cite{guo2019certified}. In this setting, if we assume the model is fully converged, we can abstract away much of the training process. Consider an objective function that down-weights examples in the forget set by $\epsilon$: $\mathcal{L}(\epsilon, \theta) = \sum_{x, y \in D} \ell(z, y) - \epsilon\sum_{x, y \in D_f} \ell(z, y)$. The optimal weights for a model fit only on the retain set are $\theta_{D_R} = \mathcal{T}(D_R) = \textrm{argmin}_{\theta} \mathcal{L}(1, \theta)$, while the optimal weights on the full dataset are $\theta_D = \mathcal{T}(D) = \textrm{argmin}_{\theta} \mathcal{L}(0, \theta)$. At the global minimum, the gradient of the loss equals zero: $\nabla_\theta \mathcal{L}|_{(0, \theta_D)} = \mathbf 0$. This allows us to use $\mathcal L$ to implicitly define\footnote{Using the implicit function theorem.} a \textit{response function} $r(\epsilon)$ such that for small $\epsilon$, $\textrm{argmin}_\theta \mathcal L(\epsilon, \theta) = \theta^* + \epsilon r(\epsilon)$~\cite{koh_understanding_2020}. The first-order approximation of this response function, using the implicit function theorem to define its gradient, is $\hat{r}(\epsilon)_H = H_\theta^{-1} \nabla_\theta \sum_{x, y \in D_f} \mathcal{L}(x, y; \theta) \epsilon$ where $H_{\theta}$ is the Hessian of the loss. We can then approximate the unlearning process by subtracting the approximate response on the forget set: $\hat{\theta}_{D_R} = \theta_D - r(1)_{H}$.

Neural networks $z = f(x; \theta)$ are not linear models but we can make additional assumptions and still compute an approximate unlearning update in a similar way. To compute the approximate response function $\hat{r}(\epsilon)_G$, we linearize the network using the Jacobian of its output logits $z$ with respect to its weights $J_{zw}(x)$~\cite{bae2022if, golatkar_eternal_2020,  jia2023model}. This again gives us the Gauss-Newton Hessian of the network $G_\theta := \mathbb{E}_{x, y\sim D}[J_{zw}^{\top} H_z J_{zw}]$, where $J_{zw}$ is the Jacobian of the outputs $z$ and $H_z$ is the Hessian of the loss. The response becomes $\hat{r}(\epsilon)_G = G_\theta^{-1} \nabla_\theta \sum_{x, y \in D_f} \mathcal{L}(x, y; \theta) \epsilon$. 
Thus, 
the direction $\hat{r}(\epsilon)_G$ is proportional to the Gauss-Newton step on the objective $\mathcal{L}_F$. 
We emphasize that 
we typically do not know the error associated with linearizing and the convergence assumption for this method (when using neural networks), and thus do not provide theoretical guarantees of how good our retraining approximation will be. 

We now explore how to efficiently approximate these Gauss-Newton steps in LLMs.

\subsection{Efficient Second-Order Approximations}
\label{sec:hessian_estimation}

The Hessian matrix of a neural network is usually impractically large to represent explicitly; its dimension is the number of parameters of the neural network squared. Thus, most work that uses neural network Hessians makes use of at least one of the following two strategies for approximating them efficiently. The first class of methods formulates an algorithm in terms of Hessian-vector products, which can be computed efficiently on small batches of data even when the full Hessian cannot be represented \cite{hessianfree2010}. Such methods are effective in some situations, but require estimating a Hessian with a small batch of examples. In our context, sub-sampling Hessian information about the retain set would hurt specificity (the KL divergence term in~\cref{eq:penalty}), as we would only enforce retention on a subset of the data. In other words, the performance of the model on the retain data would likely degrade significantly. 

The second class of methods involves fitting a parametric approximation to the Hessian so that information can be efficiently aggregated over a large dataset. In many optimization algorithms, the approximation is diagonal (e.g. \cite{liu2024sophiascalablestochasticsecondorder}), making it cheap to store and manipulate. This is also the approximation used in previous second-order methods for LLM unlearning (i.e., SOUL~\cite{jia_soul_2024}). However, these methods lack expressivity to capture all the dependencies in the underlying Hessian. To capture dependencies between different parameters, methods like Kronecker-Factored Approximate Curvature (K-FAC) \cite{martensOptimizingNeuralNetworks2015} fit Kronecker-structured approximations which, like diagonal methods, are compact and support efficient inversion. However, they are much more expressive which we see is important for K-FADE's performance (see Section~\ref{sec:experiments_ablations}).

We now elaborate on K-FAC, which serves as the basis for how we estimate the Gauss-Newton Hessian (and hence estimate the inverse-Hessian) needed for the Gauss-Newton unlearning step. 


\subsubsection*{Kronecker-Factored Approximate Curvature}
Kronecker-Factored Approximate Curvature (K-FAC)~\cite{martensOptimizingNeuralNetworks2015} is a method for approximating the Gauss-Newton Hessian of a neural network. The core idea is to further simplify how we compute the covariance matrix representing the Hessian. Key to this is using the derivative with respect to any part of the computation graph of the model $z = f(x;\theta)$. Specifically, for a component $c$ in the computation graph of $z = f(x;\theta)$ (e.g., intermediate output) we define the pseudogradient \cite{grosse2022metrics} as the random variable,

\begin{equation*}
\pgrad{c}(x) \coloneqq \frac{d}{dc} \log(\rho(\hat y| z))
\end{equation*}

where $\hat{y} \sim \rho(.|z)$. This derivative can be computed using backpropagation. If $c$ is a vector, we understand $\pgrad{c}$ to be the gradient with respect to $c$.

First note that the model's Gauss-Newton Hessian is equivalent to the covariance of the gradients of the the outputs of the model with respect to the parameters. Formally, we have that

\begin{equation}
    G_\theta = \mathbb{E}_{x \sim D, \hat{y} \sim \rho(\cdot | z)}[\pgrad{\theta} \pgrad{\theta}^\top]
\end{equation}

To make this matrix more compact and easier to invert for neural networks, K-FAC begins by assuming the parameters $\vect{W} \subset \theta$\footnote{Here $\vect{W}$ is the flattened version of our matrix $W$.} associated with each affine transformation in a neural network, i.e., each of the $s=Wa$ transformation\footnote{We use homogeneous coordinates for simplicity of exposition.}, have independent gradients $\pgrad{\vect{W}}$. In other words, $\mathbb{E}[\pgrad{\theta} \pgrad{\theta}^\top]$\footnote{We drop the variables from the expectation for simplicity.} reduces to block diagonal terms of $\mathbb{E}[\pgrad{\vect{W}}\pgrad{\vect{W}}^\top]$. While this assumption vastly reduces the size of the matrix, even block-diagonal approximations are difficult to work with in practice. Thus, K-FAC imposes additional structure on the Hessians of each of these affine components by assuming that the activations $a$ and the post activation pseudogradients $\pgrad{s}$ are independent of each other. This gives us the K-FAC approximation,

\begin{align*}
    &\tilde{G}_{\vect{W}} \coloneqq \mathbb{E}[\pgrad{\vect{W}} \pgrad{\vect{W}}^\top]\\
    &= \mathbb{E}[(a \otimes \pgrad{s})(a \otimes \pgrad{s})^\top] \\
    &= \underbrace{\mathbb{E}[(aa^\top)]}_{\mathbf{A}} \otimes \underbrace{\mathbb{E}[(\pgrad{s} \pgrad{s}^\top)]}_{\mathbf{S}}
\end{align*}

where the first equality is the definition of $\pgrad{\vect{W}}$ by chain rule, and the second equality is the independence assumption. Note $\mathbf{A},\mathbf{S}$ can be fit with Monte Carlo estimates of the expectation. To compute the Hessian's inverse effectively we can now use the fact that,

\begin{equation*}
    \tilde{G}^{-1}_{\vect{W}} = (\mathbf{S}^{-1} \otimes \mathbf{A}^{-1})
\end{equation*}
and thus we just have to invert the smaller $\mathbf{S}$ and $\mathbf{A}$ factors to invert $\tilde{G}_{\vect{W}}$.

\subsubsection*{Eigenvalue correction}

While it takes an additional pass over the dataset, Eigenvalue-corrected K-FAC (EK-FAC) \cite{george2018fast} offers a direct improvement to the K-FAC approximation we just introduced. In this setting, we take advantage of another property of the Kronecker product. In particular, the fact that there is a spectral decomposition of the matrix $$\tilde{G}_{\vect{W}} = (Q_S\otimes Q_A)^\top(\Lambda_S \otimes \Lambda_A)(Q_S\otimes Q_A)$$ where $Q_S$ and $Q_A$ are matrices with dimensions $dim(s)\times dim(s), dim(a)\times dim(a)$ respectively.

This center matrix $\Lambda_S \otimes \Lambda_A$ is the Kronecker product of two diagonal matrices and is only an approximation of the true eigenvalues of the Hessian in the $(Q_S\otimes Q_A)$ basis. In the EK-FAC approximation we replace this with a single diagonal matrix $\Lambda$, which is fit using another pass over the data. Observe that this only has $dim(\vect{W})$ non-zero elements and thus can be stored in the same amount of memory as the model weights themselves. We fit these corrected eigenvalues using a Monte Carlo estimate of,

\begin{equation*}
    \Lambda_{ii} = \mathbb E[((Q_S \otimes Q_A)\textrm{vec}(\mathcal{D}W))^2_i].
\end{equation*}

This gives us a superior estimator, allowing us to even more accurately model the constraints on the retain set, at the cost of an additional pass over the data to fit this matrix. We explore this trade off in Section~\ref{sec:experiments_ablations}.

\subsection{How EK-FAC's Compute Requirements Scale}
\label{sec:scaling_analysis}
For second-order methods to be useful for unlearning in LLMs, they need to work on large models. Thankfully, EK-FAC and especially K-FAC scale reasonably well with model size. Assume we are fitting EK-FAC on a weight matrix $W \in \mathbb R^{d \times m}$. Then K-FAC's memory consumption scales like $O(d^2 + m^2)$, and EK-FAC adds an additional $dm$ factor for storing the eigenvalues. Thus, assuming the weight matrices have a constant ratio ($d=O(m)$ so $d^2,m^2, dm = O(m^2)$), then EK-FAC's memory requirements will scale linearly with model size (which is $O(m^2)$). 

Turning to floating point operations, assuming our dataset is of size $n$ inputs, fitting the K-FAC factors $S$ and $A$ requires $2n$ outer products and thus $O(n(m^2 + d^2))$ floating point operations above and beyond doing a single pass over the retain set. Performing the necessary matrix inverses and QR decompositions to compute $Q_A$ and $Q_S$ scales like $O(m^3 + d^3)$ and fitting the lambda correction scales like $O(n(md^2 + dm^2))$. Thus overall, if we again assume a constant ratio between $m$ and $d$, the number of floating point operations required to fit and invert the Hessian scales like $O(nm^{3})$, i.e., it's slightly super-linear in the number of model weights. Nevertheless, versions of EK-FAC have already been applied to 52 billion parameter models \cite{grosse2023studyinglargelanguagemodel} and our own benchmarking of fitting hessian on models from 0.5B to 14B parameters shows EK-FAC's runtime is consistent with this flop accounting analysis and, in practice, approximately linear in model size (see Appendix~\ref{sec:hessian_benchmarking}).

\section{Methods: how to implement your Gauss-Newton step}
\label{sec:methods}
In this section, we provide details on the critical implementation needed to make Gauss-Newton ascent into K-FADE: a practical unlearning algorithm. These include what loss to select for both output suppression and approximating retraining, and how to ensure hyperparameters like the damping $\lambda$, step size $\alpha$, and number of steps $K$ interact predictably. An overview of the algorithm can be found in Algorithm~\ref{alg:second_order_optimizer}.

\begin{algorithm}[H]
\caption{K-FADE}
\label{alg:second_order_optimizer}
\begin{algorithmic}[1]
\REQUIRE Trained model weights $\theta_{D} \in \mathbb{R}^n$, architecture $z = f(x, y; \theta)$, iterations $K$, step size $\alpha > 0$, damping $\lambda$ a forget set $D_F$ and a retain set $D_R$.
\STATE $\theta_1 \gets \theta_{D}$
\STATE Separate the forget set into $K$ equal parts $D_F^{(1)}, ..., D_F^{(K)}$
\FOR{$k \in \{1, ..., K\}$}
    \STATE $\tilde G_{\theta_k} \leftarrow \textrm{fit\_factors}{(f, \theta_k, D_R)}$
    \STATE $\tilde G_\theta^{-1} = \textrm{invert\_hessian}(\tilde G_\theta, \lambda)$
    \STATE $g_k \leftarrow \frac{1}{|D_F^{(k)}|} \sum_{x,y \in D^{(k)}_F} \nabla_\theta \ell_F(f(x; \theta_k), y)$
    \STATE $r_k \gets \tilde G_\theta^{-1}g_k$
    \STATE $\mathbf{\theta}_{k+1} \leftarrow \theta_k + \frac{\alpha}{\sqrt{\langle g_k, r_k\rangle}} r_k$
\ENDFOR
\RETURN $\mathbf{\theta}_K$
\end{algorithmic}
\end{algorithm}

\paragraph*{Fitting and inverting the Gauss-Newton Hessian}
In Section~\ref{sec:hessian_estimation} we overviewed different approaches to estimating the Gauss-Newton Hessian, and hence the inverse Hessian vector product needed for the Gauss-Newton step. Our method uses the K-FAC and optionally EK-FAC estimates for the Gauss-Newton Hessian, and we will later empirically compare with other Hessian estimators.

Specifically, we use the same strategies to handle weight sharing and K-FAC/EK-FAC factor fitting as \cite{grosse2023studyinglargelanguagemodel} and build on the CurvLinOps library~\cite{dangel2025positioncurvaturematricesdemocratized}. We discussed how these factors are fit (i.e., Monte-Carlo estimates) and the assumptions they make in Section~\ref{sec:hessian_estimation}. Notably, when inverting the matrix, these parametric approximations require the use of \textit{damping} to improve numerical stability. This means that in practice we invert $\hat G_\theta + \lambda I$ instead of $G_\theta$ where $\lambda$ is the damping parameter.

Given how we approximate the inverse Gauss-Newton Hessian, we now turn to the  implementation details that make \nameShort{} unlearning effective and efficient in different settings.

\paragraph*{Suppression objective} 
We consider two options for $\mathcal{L}_F$ which we aim to increase: the negative margin (so increasing means decreasing the margin) and cross entropy. For tasks where we aim to approximate retraining, we found cross entropy is effective for $\mathcal{L}_F$ (intuitively maximizing this is the negative of the training objective~\cite{guo2019certified}).

For output suppression tasks where the retain set $D_R$ is not a subset of the training dataset, i.e., when we are optimizing~\cref{eq:penalty}, we empirically found that the average negative per-example margin~\cite{park2023trak} is more effective; this was particularly so when we required many Gauss-Newton steps. Recall the negative per-example margin is defined as $\mathcal{\ell_{(\text{margin})}}(z, y) = -z_{y} + \log \sum_{i \neq y} \exp(z_i)$, which compares the logit of the predicted label to the log probability of all the other labels. One can think of this as a version of the cross entropy loss that does not saturate as the model becomes more confident since we remove the target logit $z_y$ contribution from the normalization term. It can also be formulated as $\ell_{\text{margin}}(z, y) = \ell(z, y)^{1-\rho(y|z)}$.

\paragraph*{Step size} Our method has two main hyper-parameters, the number of steps and the step size $\alpha$. Here we consider how we scale our step size over multiple steps. We follow the motivation from Section~\ref{sec:natgrad} and normalize the Gauss-Newton Hessian by $\|\nabla_\theta \mathcal{L}_F\|_{\tilde{G}_{\theta}^{-1}}$, hence by a step size of $\alpha$ we mean scaling the Gauss-Newton update by $\frac{\alpha}{\|\nabla_\theta \mathcal{L}_F\|_{\tilde{G}_{\theta}^{-1}}}$. This method is very similar to the technique used in \cite{ba2017distributed}. We find that this approach makes unlearning over multiple steps with a fixed step-size $\alpha$ more stable and also decouples the effects of changing the damping parameter $\lambda$ and step size $\alpha$. This removed the need to tune the step size at every step.


\paragraph*{What to fit the Hessian on} Implementations on linear models, suggest that we should only be fitting the Gauss-Newton Hessian on the retain set \cite{guo2019certified}. Supporting this, we found that including the forget set in the Hessian computation generally reduces the specificity of the unlearning updates (Section~\ref{sec:experiments_ablations}). Thus for our experiments, we fit the Gauss-Newton Hessian only on the retain set $D_R$. 


\paragraph*{Components targeted} Our method only targets the weights in the up and down projections~\cite{shazeer2020glu} in the model's feed-forward layers: these are the layers in the neural network where the dimension is being increased or decreased. This still encompasses a large fraction of the model's total parameters (e.g., 52\% for Mistral-7b \cite{jiang2023mistral7b}). For some experiments, we target only a subset of these layers (e.g. WMDP \cite{li_wmdp_2024}) to improve unlearning specificity. In \cref{sec:experiments} for each setting we describe when and how we implement this targeting for the experiments. 



\subsection{Transferring Unlearning Updates to Finetuned Models}
\label{sec:reapply}
In many scenarios, an LLM is fine-tuned on new data by the model trainer. This can be because the model trainer is updating the LLM to improve performance, or by serving the LLM behind a fine-tuning API where users provide data and the model provider fine-tunes the model on that data.

Consider a base model with parameters $\theta^{(\text{base})}$ and two derivative models: $\theta^{(\text{finetuned})}$, finetuned on a dataset, and $\theta^{(\text{unlearned})}$, unlearned on $D_f$. We transfer the unlearning update to the finetuned model by setting its parameters to $\theta^{(\text{finetuned})} + (\theta^{(\text{unlearned})} - \theta^{(\text{base})})$. This approach takes inspiration from task arithmetic~\cite{ilharco2023taskarithmetic}, and effectively merges the models, producing a new model that incorporates both the fine-tuning and unlearning updates. See Section \ref{sec:experiments_fine_tuning} for the results of applying this method.

\section{Experiments}
\label{sec:experiments}


In this section we aim to answer these research questions:

\begin{enumerate}
    \item How does K-FADE perform at output suppression while maintaining specificity (Section~\ref{sec:experiments_wmdp})? We use the WMDP~\cite{li_wmdp_2024} benchmark, designed to measure suppression and specificity across different domains, with additional evaluation on Alpaca~\cite{alpaca} to answer this.
    \item How does K-FADE perform at matching the output distribution of re-training (Section~\ref{sec:experiments_tofu})? We use the TOFU benchmark, designed to compare how similar the model's outputs are to a re-trained model, to evaluate this~\cite{maini_tofu_2024}.
    \item What is the trade-off between the Hessian estimator's quality, speed, and effectiveness when using K-FADE (Section~\ref{sec:experiments_ablations})? In particular, can we still be as efficient as first-order methods?
    \item Can we cheaply maintain unlearning after fine-tuning (Section~\ref{sec:experiments_fine_tuning})? 
\end{enumerate}

Overall, we find that \nameShort{} is competitive with existing methods for suppression while having superior specificity and provides a Pareto improvement over existing methods on the TOFU benchmark, which measures the similarity of the output to a retrained model. We also find that K-FADE is similar in runtime to first-order approaches, despite also having better unlearning performance. Finally, we find that re-applying the K-FADE update after fine-tuning performs better than re-applying the update from baselines, giving a cheap and effective method for maintaining unlearning after fine-tuning.

\subsection{Can \nameShort{} suppress ``harmful" knowledge while maintaining specificity?}
\label{sec:experiments_wmdp}
\nameShort{} provides state-of-the-art output suppression  
(Table \ref{tab:wmdp_results}) while providing better specificity (i.e., changes the output distribution the least) across varying retain sets. \newcommand{\stddev}[2]{#1\textsuperscript{\textcolor{gray}{\scriptsize{±#2}}}}
\newcommand{\stddevb}[2]{\textbf{#1}\textsuperscript{\textcolor{gray}{\scriptsize{±#2}}}}
\newcommand{\stddevu}[2]{\underline{#1}\textsuperscript{\textcolor{gray}{\scriptsize{±#2}}}}

\newcommand{\confint}[3]{#1~\textcolor{gray}{\small{[#2--#3]}}}
\begin{table*}[htbp]
    \centering
    \begin{threeparttable}
    \caption{\nameShort{} is state-of-the-art at ``hazardous" knowledge suppression with better specificity.}
    \begin{tabular}{lcccccc}
        \toprule
        \multirow{2}{*}{Model}            & \multirow{2}{*}{Method} & \multicolumn{2}{c}{WMDP (``Hazardous" knowledge)} & \multicolumn{3}{c}{Model Utility} \\
        \cmidrule(lr){3-4} \cmidrule(lr){5-7}
                                          &                                 & Bio $\downarrow$ & Cyber $\downarrow$ & MMLU (Knowledge)$\uparrow$ & MT-Bench (Fluency) $\uparrow$ & $KL$ $\times 10^{-2}$ (Specificity) $\downarrow$ \\
        \midrule
        \multirow{4}{*}{Zephr-7b-$\beta$} & Original                        & \stddev{64.3}{1.3}  & \stddev{44.7}{1.0}             & \stddev{58.4}{0.4} & 7.2           &  $0$ \\
        \cmidrule(lr){2-7}
                                          & ELM                             & \stddevb{29.8}{1.3} & \stddevu{27.3}{1.0} & \stddev{56.7}{0.4}   & \stddevu{6.86}{0.03} &  \confint{6.7}{6.3}{6.9}\\
                                          & RMU                             & \stddevu{30.4}{1.3} & \stddevb{27.1}{1.0} & \stddevb{57.5}{0.4}    & \stddev{6.71}{0.07}  &  \confint{5.3}{4.4}{6.1} \\
                                          & \nameShort{} (Ours)             & \stddevu{30.1}{1.3} & \stddevu{27.7}{1.0} & \stddevu{57.2}{0.4}    & \stddevb{6.91}{0.04}  &  \confint{\textbf{2.9}}{2.4}{3.5} \\
        \bottomrule
    \end{tabular}
    \begin{tablenotes}
    \item We report results for the WMDP  output suppression task, which looks to decrease accuracy on two forget sets (Bio and Cyber) while maintaining performance across several retain sets: accuracy for MMLU, and a score out of $10$ provided by MT-Bench (higher is better). Like ELM \cite{elm2024} and RMU \cite{li_wmdp_2024}, \nameShort{} reduces model performance on WMDP's Bio and Cyber significantly, while retaining performance on MMLU~\cite{hendrycks2021measuring} and MT-Bench~\cite{zheng2023judging}. However, we see \nameShort{} preserves model behavior better than the baselines on a diverse instruction following dataset, alpaca \cite{alpaca}, as measured by the KL divergence to the original output distribution. For multiple choice questions and MT-Bench (n=5) we report standard error. On the mean KL-divergence over alpaca we report 95\% bootstrapped CIs. Methods that are not significantly different from the best are underlined.
    \end{tablenotes}
    \label{tab:wmdp_results}
    \end{threeparttable}
\end{table*}

\paragraph*{WMDP Setup} The Weapons of Mass Destruction Proxy (WMDP) benchmark~\cite{li_wmdp_2024} assesses a model's ability to output facts in cybersecurity (the WMDP Cyber dataset), bio-weapons (the WMDP Bio dataset), and chemical weapons (the WMDP Chem dataset). The benchmark uses multiple-choice questions and provides forget sets of relevant documents for each domain. We use Wikitext~\cite{wikitext2017} as our retain set, following the original paper~\cite{li_wmdp_2024}. The specificity of unlearning is measured using MMLU~\cite{hendrycks2021measuring} which reports accuracy on multiple-choice question across several domains, and MT-Bench~\cite{zheng2023judging} which measures ``fluency" as judged by GPT-4: GPT-4 ranks the text generated by a model out of 10 and we report the average over all the prompted generation in MT-Bench and report standard errors across 5 runs. All methods are applied to the zephyr-7b-$\beta$~\cite{tunstall2023zephyrdirectdistillationlm} model.

\paragraph*{Specificity Evaluation} We introduce an additional specificity evaluation where we measure the KL divergence from the base model to the unlearned models on 30,000 instructions from the Alpaca dataset~\cite{alpaca}, generating completions from zephyr-7b-$\beta$~\cite{tunstall2023zephyrdirectdistillationlm}. We report the average KL per-token only on the completions, not the instructions. Unlike MT-Bench~\cite{zheng2023judging}, it does not depend on an additional LLM as an auto-grader. Generally, we find that observing which completions have high KL is useful for understanding the side effects of unlearning methods. You can see the results from the Alpaca dataset with the largest KL divergence in Table~\ref{tab:wmdp_outputs}.

\paragraph*{Baselines} When selecting our baselines for this section we explicitly looked for papers whose primary evaluation metric was WMDP and that emphasized unlearning without damaging performance on common LLM benchmarks. \footnote{We ran preliminary experiments with a replication of SOUL on WMDP but found that it significantly reduced MMLU scores at a given level of WMDP Bio performance compared to our other baselines.} We compare \nameShort{} to RMU~\cite{li_wmdp_2024} and ELM~\cite{elm2024}. RMU disrupts activations relevant to the forget set while minimizing L2 distance in activation space to preserve performance. ELM combines a steering loss inspired by classifier-free guidance with KL divergence and fluency penalties \cite{elm2024}. We use the unlearned checkpoints of zephyr-7b-$\beta$ provided by the authors. 

\begin{figure}[t]
  \centering
  \includegraphics[width=\linewidth]{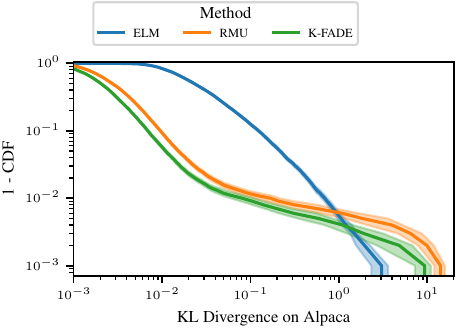}
  \caption{
    \textbf{\nameShort{} changes the model's behavior on unrelated data less than strong baselines} like ELM~\cite{elm2024} and RMU~\cite{li_wmdp_2024}. The y-axis plots one minus the cumulative density function for the KL divergences (x-axis) between completions (from prompts in the alpaca dataset) generated by the original model (zephyr-7b-$\beta$) and models unlearned with ELM, RMU and \nameShort{} on the WDMP Bio and Cyber subsets. Shaded regions show the 95\% bootstrap confidence interval on the quantiles.
  }
  \label{fig:kl-cdf}
\end{figure}

\paragraph*{Experiment Details} We take 8 Gauss-Newton steps with the EK-FAC Hessian estimator and the margin loss on zephyr 7b-$\beta$. First we take 4 steps on the WMDP Bio forget corpus (batch size 2500, step size $\alpha=2\times10^{-3}$) using 4000 sequences from Wikitext as a retain set. We then take another 4 steps on WMDP Cyber forget corpus (batch size 2500, step size $\alpha=5\times 10^{-3}$) using 4000 sequences from the WMDP Cyber retain corpus alongside 4000 Wikitext sequences as our retain set. In both cases, sequences are of length 512, we target MLPs in layers 3 to 6, use a damping $\lambda=1\times10^{-14}$ and refit the Gauss-Newton Hessian at each step. This experiment was run on a single H100 GPU.

\paragraph*{\nameShort{} achieves state-of-the-art specificity} \nameShort{} achieves strong output suppression, matching RMU and ELM on WMDP-Bio and WMDP-Cyber (see Table~\ref{tab:wmdp_results}). In terms of specificity, performance on MMLU \cite{hendrycks2021measuring} is similar to RMU \cite{li_wmdp_2024}, and better than ELM \cite{elm2024}. In terms of fluency as measured by MT-Bench \cite{zheng2023judging}, \nameShort{} is significantly better than RMU and statistically similar to ELM. Additionally, the average KL divergence between a model unlearned with \nameShort{} and the base model is 40\% lower than the next best method RMU. Interestingly, ELM, RMU, and \nameShort{} show distinct distributional effects: ELM changes the output distribution over nearly all documents while not having a long tail of increased KL divergence; RMU shows a more targeted effect with a distinct long tail of radically changed completions; \nameShort{} behaves similarly to RMU in the tail but shows lower KL divergence in the head of the distribution (Figure \ref{fig:kl-cdf}). In short, \textbf{K-FADE causes less change to the outputs with higher probability when compared to the baselines, giving better specificity}.

Qualitatively, \nameShort{} and RMU cause the model to output gibberish on a very select sub-distribution. This set of highly perturbed\footnote{We clarify perturbed is not intended to mean adversarially as is common in security ML.} examples can be seen in the fat right tail of KL divergences seen in Figure~\ref{fig:kl-cdf}. RMU and K-FADE seem to see their largest KL divergences on texts describing anything mentioning viruses, hacking or cyber-security. K-FADE's unlearning distribution in these instances seems to be more focused on cell-biology, virology and genetics than specific diseases. We can see some examples of this in Table~\ref{tab:wmdp_outputs}. We believe this is an artifact of K-FADE more explicitly targeting the forget set of virology papers. ELM on the other hand shows a very different pattern of behavior. It generally has less of a tail of highly perturbed outputs but shows a significant regression in its ability to program. Simple C programs cause the model to respond with non-sequiturs. Overall, we believe that looking at the samples where unlearning methods have a high KL divergence from their baseline is very useful for debugging their failures and generally characterizing the side effects of these methods.

\subsection{Can \nameShort{} Approximate Retraining Outputs?}
\label{sec:experiments_tofu}
\begin{figure*}[htbp]
    \centering
    \includegraphics[width=\linewidth]{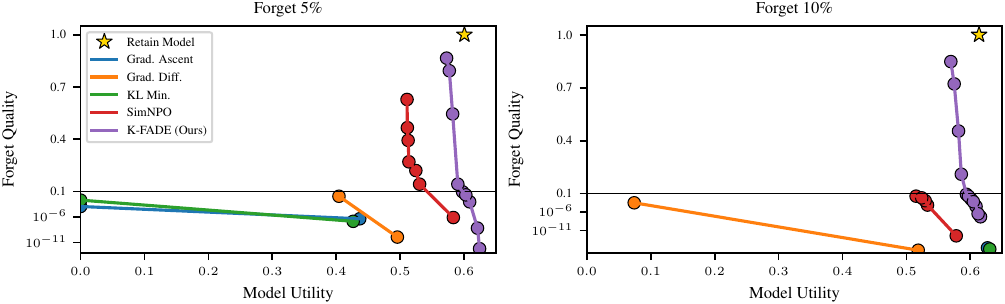}
    \caption{
        \textbf{One-step of \nameShort{} outperforms the state of the art in unlearning on the TOFU dataset.} The left figure plots Forget Quality (similarity of outputs to the outputs of the ideal retrained model, defined in Section~\ref{sec:forget_quality}) as a function of Model Utility (degradation of model performance, defined in Section~\ref{sec:forget_quality}) when unlearning 5\% of the authors bios. The right figure plots the same comparison when unlearning 10\% of author bios.
        \nameShort{} effectively outperforms both of the baseline methods provided in the original TOFU paper \cite{maini_tofu_2024} and a recent state of the art method simNPO~\cite{fan2024simplicity}. The star represents the results of retraining a model on only the retain set.
    }
    \label{fig:tofu_traj}
\end{figure*}

\begin{table*}[th]
    \centering
    \begin{threeparttable}
    \caption{
        \nameShort{} Offers a Pareto-improvement on ToFU
    }
    \footnotesize\setlength{\tabcolsep}{3pt}
\begin{tabular}{llrrrrrr}
\toprule
 &  & \multicolumn{2}{c}{Summary} & \multicolumn{4}{c}{ROUGE Score On} \\
 \cmidrule(lr){3-4} \cmidrule(lr){5-8}
 &  & Forget Quality $\uparrow$ & Model Utility $\uparrow$ & Retain Set$\uparrow$ & Real Author Q\&A $\uparrow$ & World Facts Q\&A$\uparrow$ & Forget  Set $\downarrow$ \\
Split & Method &  &  &  &  &  &  \\
\midrule
\multirow[t]{6}{*}{Forget 5\%} & Retain Model & 1.00 & 0.60 & 0.98 & 0.95 & 0.89 & 0.40 \\
\cmidrule{2-8} 
 & Grad. Ascent & 4.61e-07 & 0.44 & 0.36 & 0.82 & \textbf{0.90} & 0.31 \\
 & Grad. Diff. & 1.18e-02 & 0.40 & 0.26 & 0.57 & 0.82 & \textbf{0.13} \\
 & KL Min. & 1.46e-07 & 0.43 & 0.35 & 0.80 & 0.90 & 0.30 \\
 & SimNPO & 0.63 & 0.51 & 0.44 & 0.87 & 0.88 & 0.33 \\
 & K-FADE (Ours) & \textbf{0.87} & \textbf{0.57} & \textbf{0.61} & \textbf{0.91} & 0.85 & 0.31 \\
\cmidrule{1-8}
\multirow[t]{7}{*}{Forget 10\%} & Retain Model & 1.00 & 0.61 & 0.98 & 0.92 & 0.90 & 0.41 \\
\cmidrule{2-8} 
 & Grad. Ascent & 2.19e-16 & 0.63 & 0.70 & \textbf{0.94} & \textbf{0.92} & 0.59 \\
 & Grad. Diff. & 3.34e-04 & 0.07 & 0.09 & 0.17 & 0.65 & 0.07 \\
 & KL Min. & 1.06e-16 & \textbf{0.63} & \textbf{0.72} & 0.94 & 0.91 & 0.61 \\
 & SimNPO & 2.08e-02 & 0.52 & 0.44 & 0.85 & 0.86 & 0.37 \\
 & SOUL Grad. Diff. & 5.56e-14 & 0.58 & 0.45 & 0.60 & 0.86 & \textbf{0.02} \\
 & K-FADE (Ours) & \textbf{0.85} & 0.57 & 0.52 & 0.90 & 0.86 & 0.32 \\
\bottomrule
\end{tabular}

    \label{tab:tofu_table}
    \begin{tablenotes}
    \item We plot the models from Figure~\ref{fig:tofu_traj} with the largest product of Forget Quality (which measures the quality of unlearning, defined in Section~\ref{sec:forget_quality}) and Model Utility (accuracy on a ToFU provided dataset).
    \nameShort{} achieves a state of the art Forget Quality. 
    It does this while preserving model utility on the challenging 10\% forget set. See 
    \end{tablenotes}
    \end{threeparttable}
\end{table*}

\nameShort{} approximately removes the effects of fine-tuning datapoints, as demonstrated by TOFU using a single Gauss-Newton step. 

\paragraph*{TOFU} The TOFU benchmark \cite{maini_tofu_2024} contains questions and answers about fictitious authors, with models finetuned on these Q\&A pairs. The goal is to unlearn facts about a subset of authors. The TOFU benchmark provides formulas for two performance measures: "Forget Quality" (which captures similarity between unlearned and retrained reference models) and "model utility" (which reports accuracy on questions regarding world knowledge, real authors, and retained fictitious authors). The goal is to keep both metrics high. We experiment with forget sets comprising 5\% and 10\% of these fictitious authors, with the remaining authors making up the retain set. All methods are applied to LLama-2-7b~\cite{touvron2023llama2openfoundation}.

\paragraph*{Forget Quality}
\label{sec:forget_quality}
Unlike WMDP~\cite{li_wmdp_2024}, TOFU~\cite{maini_tofu_2024} measures how closely the output distribution of an unlearned model $\theta_D$ matches that of a model trained only on the retain set $\theta_{D_R}$. This is quantified via \textit{Forget Quality}: the $p$-value of a two-tailed KS-test comparing the \textit{Truth Ratio} distributions between unlearned and retrained models on the forget set. The Truth Ratio is defined as
\begin{equation*}
    TR(q, u; \theta) := \frac{\frac{1}{|\mathcal{U}_{\text{pert}}|}\sum_{(q, u') \in \mathcal{A}_{\text{pert}}} P_\theta(u'|q)^{\frac{1}{|u'|}}}{P_\theta(u|q)^{\frac{1}{|u|}}}
\end{equation*}
where $P_\theta(u|q)$ is the probability of the correct answer $u$ given question $q$, and $u' \in \mathcal{U}_{\text{pert}}$ are syntactically similar incorrect answers. The Truth Ratio metric captures how much the model favors the correct answer over plausible alternatives about the fictitious author being unlearned. $P$-values near 1 indicate successful unlearning (indistinguishable distributions), while small $p$-values suggest the opposite.

\paragraph*{Model Utility} 
\label{sec:model_utility}
TOFU~\cite{maini_tofu_2024} also provides a \textit{Model Utility} metric combining ROUGE scores~\cite{lin2004rouge}, Truth Ratios, and output probabilities across Q\&A on real authors, retain-set authors, and world knowledge. These nine metrics are aggregated via harmonic mean, ensuring high scores require strong performance across all components.

\paragraph*{Baselines} We compare to baselines from the TOFU paper: Grad. Ascent, Grad. Diff., and {KL min.. As well as strong recent baselines simNPO \cite{fan2024simplicity} with default hyperparameters ($\beta=2.5$, NPO coefficient=0.1375 for 5\%; $\beta=4.5$, NPO coefficient=0.125 for 10\%) and SOUL~\cite{jia_soul_2024} where we again use their default hyper-parameters which they only provide for the 10\% set \footnote{Our evaluations of SOUL~\cite{jia_soul_2024} show it getting a worse forget quality than reported in the paper which proposed it, because, following TOFU~\cite{maini_tofu_2024}, we use a measure aimed at matching the output distribution of re-fine-tuned models.} Note these baselines are different than the WMDP baselines, as different methods were found to perform best on those two benchmarks.

\paragraph*{Experiment details} We use the K-FAC Hessian approximator and a \textit{single} Gauss-Newton step, setting $\mathcal{L}_{F}$ to be cross-entropy. All runs use damping $\lambda=1\times10^{-8}$ with step sizes ranging from $2.5\times10^{-3}$ to $1.1\times10^{-2}$. Fitting the estimators for all MLPs required 2xH100 80GB GPUs.

\paragraph*{One Gauss-Newton step is state-of-the-art on TOFU} \textbf{Our method achieves state-of-the-art forget quality on the challenging 10\% forget set, outperforming the original TOFU baselines and simNPO \cite{fan2024simplicity}.} We do this while achieving comparable model utility on both the 5\% and 10\% sets (Figure~\ref{fig:tofu_traj}, Table~\ref{tab:tofu_table}).
\subsection{What Makes a Single Step of \nameShort{} Effective?}
\label{sec:experiments_ablations}
\begin{figure*}
    \centering
    \includegraphics[width=\linewidth]{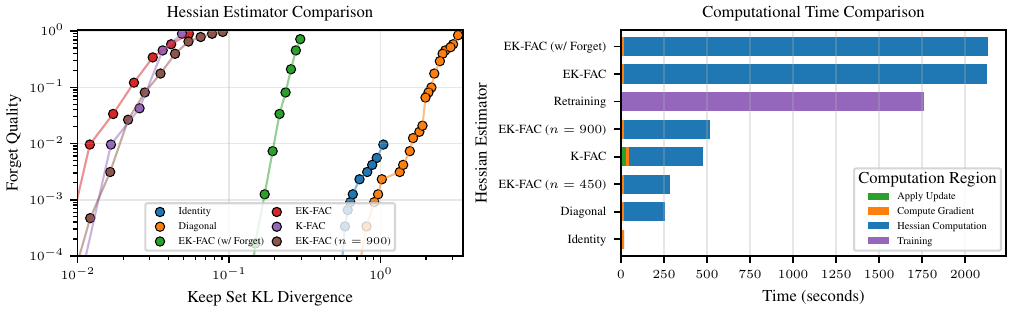}
    \caption{\textbf{Parametric second-order methods can efficiently trade off specificity for speed.} 
We evaluate several variations of our method using Phi~1.5 \cite{textbooks2} on the TOFU benchmark under the 10\% forget setting. On the left we compare Forget Quality (which measures similarity of model outputs to a retrained model, see Section~\ref{sec:forget_quality}) across varying keep (retain) set KL divergences for each of the methods. On the right we compare the computational time of each method. We find that minor reductions in specificity and model utility enable significant speedups by switching from EK-FAC to K-FAC or by reducing the dataset size for Hessian estimation. Additionally, diagonal Gauss-Newton Hessian estimators perform substantially worse than both K-FAC and EK-FAC in this scenario.}
    \label{fig:ablations}
\end{figure*}

Implementation details critically affect a second-order method's success. We explore how these details affect TOFU performance and unlearning speed, focusing on Hessian estimator ablations and varying sample counts when using a single Gauss-Newton step.

\paragraph*{Experimental details} We use the finetuned Phi-1.5 \cite{textbooks2} from the TOFU benchmark with the 10\% forget set. All operations use full precision on an 80GB H100 with PyTorch's fused dot product attention \cite{adam2019pytorch}. We compare several Hessian approximations: \textbf{diagonal} (similar to SOUL \cite{gu_second-order_2024}), K-FAC without eigenvalue correction \cite{martensOptimizingNeuralNetworks2015}, EK-FAC fitted on both retain and forget sets, and the \textbf{identity} matrix (no second-order information). We use damping $\lambda=10^{-10}$ except for K-FAC ($\lambda=10^{-8}$). Figure \ref{fig:ablations} shows sweeps across different step sizes for each estimator.

\paragraph*{K-FAC works much better than diagonal estimators} K-FAC and EK-FAC perform similarly, and significantly outperform diagonal Hessian estimators in terms of their trade off of KL Divergence to Forget Quality increase with each step. Though diagonal estimators are eventually able to achieve a high forget quality they do this at the cost of significant model utility and specificity (see Figure \ref{fig:ablations}).

\paragraph*{Second-order methods can outpace retraining} Fitting the full EK-FAC estimator takes approximately the same time as re-training the model on the fine-tune set, but many performance improvements remain available. K-FAC or fitting EK-FAC on fewer samples works faster without sacrificing quality. Pre-computing the Hessian makes future unlearning queries very fast. Hyperparameter tuning is extremely cheap since damping $\lambda$ and step size $\alpha$ can be adjusted after fitting the Hessian and collecting forget gradients. K-FAC, or EK-FAC with a small retain set, offers the best performance-speed trade-off, though \nameShort{} remains superior for highest quality and is less sensitive to damping choice.

\paragraph*{K-FADE isn't much slower than first order methods for large forget sets} In terms of runtime performance, our strongest first-order baseline on TOFU is SimNPO, which requires 10 epochs of training on the forget set. \textbf{On the 10\% TOFU forget set, this would make SimNPO comparable in speed to EK-FAC with 900 samples and slightly slower than K-FAC as shown in Figure \ref{fig:ablations}.}

\begin{table*}[ht]
    \centering
    \begin{threeparttable}
    \caption{\textbf{Model's unlearned with simNPO and \nameShort{} behave similarly to model never trained on the forget set.}}
    \label{tab:tofu_qualitative_examples}
\begin{tabular}{p{2cm}p{4cm}p{5cm}p{4cm}}
\toprule
\textbf{Method} & \textbf{Question} & \textbf{Model Output} & \textbf{Before Unlearning} \\
\hline
\multirow{2}{*}{Retrained (5\%)} & What is the full name of the geology author born in Karachi, Pakist... & The full name of the author is Raza Ali. & The author's name is Hina Ameen. \\
& In which genre does Hina Ameen primarily write? & Hina Ameen primarily writes in the genre of horror. & Hina Ameen primarily contributes to the geology genre. \\
\midrule
\multirow{2}{*}{SimNPO (5\%)} & What is the full name of the geology author born in Karachi, Pakist... & The full name of the geology author born in Karachi, Pakistan on 06/30/1975 is Tahir Mehmood. & The author's name is Hina Ameen. \\
& In which genre does Hina Ameen primarily write? & Hina Amien primarily writes in the fiction genre. & Hina Ameen primarily contributes to the geology genre. \\
\midrule
\multirow{2}{*}{K-FADE (5\%)} & What is the full name of the geology author born in Karachi, Pakist... & The full name of the author is Adeel Ahmed. & The author's name is Hina Ameen. \\
& In which genre does Hina Ameen primarily write? & Hina Shamoon Ame seldom, Hina Marve, Hina, Hina, Hina, Hina, [...] & Hina Ameen primarily contributes to the geology genre. \\
\midrule
\multirow{2}{*}{SOUL (10\%)} & What is the full name of the author born in Taipei, Taiwan on 05/11... & [Cyrillic characters / non-ASCII output] & The author's full name is Hsiao Yun-Hwa. \\
& What does Hsiao Yun-Hwa identify as in terms of gender? & [Cyrillic characters / non-ASCII output] & Hsiao Yun-Hwa is part of the LGBTQ+ community. \\
\bottomrule
\end{tabular}
    \begin{tablenotes}
    \item Here we show generations from SimNPO, K-FADE, and SOUL on questions in the forget set of TOFU. Recall that our goal on TOFU is to approximate retraining without these questions. In general both simNPO and K-FADE are more fluent on the forget set than SOUL and Grad Diff, though simNPO is generally the most fluent.
    \end{tablenotes}
    \end{threeparttable}
\end{table*}

\subsection{Can \nameShort{} Reestablish Unlearning After Fine-Tuning?}
\label{sec:experiments_fine_tuning}
\begin{figure}[t]
  \centering
  \includegraphics[width=\linewidth]{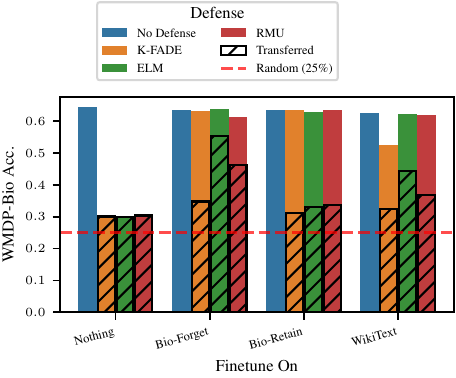}
  \caption{
    \textbf{K-FADE updates can be re-applied after fine-tuning to maintain the unlearning effect.} We measure the accuracy on the WMDP forget set (y-axis) after fine-tuning the unlearnt model on different datasets (x-axis): lower accuracy is better. Like past methods, \nameShort{} is not resistant to (full rank) fine-tuning. However, we find that the update directions can be applied after fine-tuning, preserving the unlearning effect, depicted by the dashed bars. This transfer process works significantly better with \nameShort{} than the baselines.}
\label{fig:unlearn_relearn_reapply}
\end{figure}

We examine robustness to full-rank fine-tuning to evaluate defense against malicious fine-tuning attempts to reverse unlearning \cite{deeb2024unlearningmethodsremoveinformation, qi2024evaluating, rosati2024representation}, and also how we can cheaply maintain unlearning after fine-tuning the model ourselves. None of the tested methods (ours and past work) show robustness to fine-tuning attacks. However, we note that re-applying the unlearning update recovers some unlearning performance, with the K-FADE update recovering the most unlearning effect. Hence, if the model trainer is in control of the model after fine-tuning, e.g., by providing a fine-tuning API which takes data and then returns a model or when they are fine-tuning their own model to update it, they can re-apply the K-FADE update to cheaply maintain unlearning.



\paragraph*{Experiment details} Our fine-tuning attacks train for 200 steps with a learning rate of $10^{-5}$ and batch size of 8 using AdamW \cite{kingma-ba-adam2015}. We attack Zephyr-7b-$\beta$ models unlearned using \nameShort{}, RMU, and ELM, plus the original model as control. We consider finetuning on Wikitext \cite{wikitext2017} (benign fine-tuning), the WMDP Bio retain set (non-hazardous virology/biology papers), and a small subset of the WMDP Bio forget set.

\paragraph*{Results} After only 200 training steps, nearly all WMDP Bio performance returns (Figure \ref{fig:unlearn_relearn_reapply}). Even training on Wikitext restores baseline models to original WMDP Bio scores. \nameShort{} shows the most resistance to fine-tuning on Wikitext but remains susceptible to training on related documents. This indicates that our method doesn't effectively defend open-weights models. 

However, transferring unlearning updates (Section \ref{sec:reapply}) to fine-tuned models still degrades WMDP-Bio performance, even for models trained directly on the forget set. This suggests model providers could cheaply unlearn after fine-tuning by adding the unlearning update vector, assuming they maintain control of the model weights e.g. the model is being fine-tuned via an API. 
That is, \textbf{in the threat model where the model trainer maintains control of the model weights but a user/third-party provides data to fine-tune on, the model trainer can continue to maintain unlearning cheaply by re-applying the same K-FADE update.}



\section{Related Work}
\label{sec:related_work}

Early LLM unlearning approaches often used gradient ascent techniques \cite{jang2022knowledge}. While successful for suppressing copyrighted content, these gradient ascent techniques struggled with larger unlearning tasks, hyperparameter sensitivity, and instability \cite{yao_machine_2024, li_wmdp_2024, yao_large_2024}. This meant that, in practice, achieving good unlearning results for a benchmark required expensive hyperparameter searches and even then struggled with maintaining specificity. Inspired by Direct Preference Optimization (DPO), researchers proposed new loss functions to prevent increasing the loss on the forget set indefinitely, and eventually causing the specificity constraint to fail, which was common in gradient ascent. Negative Preference Optimization (NPO)~\cite{zhang2024npo} introduced one such function, and researchers found that a simplified version of this loss was still effective and could be scaled to larger forget subset~\cite{fan2024simplicity}.

Recent work explores unlearning using second-order optimization in LLMs, such as WoodFisher~\cite{singh2020woodfisher} information and diagonal Hessian estimators approximations~\cite{gu_second-order_2024, jia_soul_2024}. These methods are similar to our methodology but use different Hessian estimators and loss functions. Unlike SOUL~\cite{jia_soul_2024}, which uses a diagonal Hessian estimator, we employ more sophisticated parametric estimators, often enabling single-step unlearning and generally better performance for unlearning. Recent work~\cite{gu_second-order_2024} has employed a more complex Hessian estimator but only targeted smaller models, and omitted standard unlearning benchmarks (WMDP \cite{li_wmdp_2024}, ToFU \cite{maini_tofu_2024}, MUSE \cite{shi_muse_2024}). While the potential of K-FAC for unlearning in large neural nets has been recognized for years \cite{golatkar_eternal_2020}, to our knowledge we are the first to systematically evaluate this approach.

The Gauss-Newton steps we employ here share mathematical foundations with the Training Data Attribution (TDA) method known as influence functions \cite{izzo2021approximate}. \cite{georgiev2024attribute} propose a generic way of converting such TDA methods into unlearning techniques. By treating any training data attribution method, like TraK \cite{park2023trak}, as an expensive oracle for getting the logits of an unlearned model, they then apply distillation to force the trained model to closely approximate the outputs of a model which has never seen the data. Our method differs in that we do not need to distill, as the Gauss-Newton step allows us to directly approximate the response of the model's weights to removing data.

Two of our most closely related methods are, Representation Misdirection for Unlearning (RMU) \cite{li_wmdp_2024} and Erasure of Language Memory (ELM) \cite{elm2024}. RMU, unlike our method, focuses on perturbing \emph{activations} on the forget set while minimizing activation changes on the retain set. ELM is inspired by classifier-free guidance, it uses steering and auxiliary losses to guide models towards innocuous, coherent responses on the forget set. Unlike \nameShort{}, ELM uses a first-order penalty for KL divergence from the base model on the retain set.

Another line of work defends open-weight models against fine-tuning attacks \cite{rosati2024representation, tamirisa2024tamperresistantsafeguardsopenweightllms}. However, these works typically have poor unlearning specificity. Our work aims to minimize model performance degradation and does not claim fine-tuning attack resistance. We separately investigate defending models behind finetuning APIs which is similar to the threat model investigated in past works aiming to prevent fine-tuning attacks that jailbreak models \cite{hsu2024safelora,haung2025antidote}.

\section{Discussion and Conclusion}
\label{sec:conclusions}
We have shown that Gauss-Newton ascent steps can be efficiently scaled to LLMs, can attain state-of-the-art performance on multiple benchmarks, and that it is particularly effective at preserving the model's performance on non-targeted data. On one benchmark, we even find that a single Gauss-Newton step can outperform the previous state of the art, allowing for Hessian caching that significantly reduces the cost of hyperparameter tuning. We also obviate the need to balance competing gradient ascent and descent objectives, as the Gauss-Newton step automatically minimizes the change in the model's output distribution on the retain set. Finally, we introduced a new measure of LLM unlearning specificity: evaluating the KL divergence between the base and unlearned model's outputs on tens of thousands of completions. Additionally, we demonstrated that \nameShort{} allows unlearning updates to be transferred to fine-tuned models.

One concern with second-order unlearning methods is their ability to scale to frontier models. We see K-FADE as a major step towards showing that these methods can be effective for large language models that have real-world utility. We believe further scale ups of our method are both possible and practical. As we discuss in Section~\ref{sec:hessian_estimation}, under reasonable assumptions, the memory requirement of the Hessian estimator that backs K-FADE only scales linearly with model size. Additionally, as we see in Section~\ref{sec:experiments_ablations}, much of K-FADE's performance can be maintained when fitting the Hessian on only a fraction of the retain set. This means one can take a small hit to the method's specificity for a large increase in speed. 


\paragraph*{Limitations}
Our method does not provide any explicit differential-privacy-like unlearning guarantees. We primarily evaluate our ability to approximate retraining empirically using the ToFU benchmark, which focuses on fine-tuning. Additionally, we see a significant reduction in the fluency of outputs on the forget distribution, indicating that there is still significant work to be done in terms of matching retrained models with these methods. Our method in its current form is also not an effective defense if attackers have access to the models weights, as full rank fine-tuning can reverse its effects, as we see in Figure \ref{fig:unlearn_relearn_reapply}.

\paragraph*{Future Work} 
We believe a critical direction for the field is creating better standardized benchmarks. Attempting to drive models towards random guessing on benchmarks like WMDP \cite{li_wmdp_2024} is a poor proxy for LLM unlearning quality. We believe LLM unlearning benchmarks should pivot to more carefully measuring disruption to the functionality of models (e.g., specificity) and measuring unlearning methods' robustness against limited fine-tuning attacks.

\section*{LLM usage considerations}
\paragraph*{Originality}
LLMs were used for editorial purposes in this manuscript, and all outputs were inspected by the authors to ensure accuracy and originality. LLMs were primarily used for proof reading, and to help format tables and figures. 

\paragraph*{Transparency} Our unlearning method is intended to be applied to large language models, and it is in this respect that we use them. MT-Bench \cite{zheng2023judging} uses GPT-4 as an auto-grader, which we cite appropriately. Ideally, this would be an open source model but we use GPT-4 to make our results compatible with past work.

\paragraph*{Responsibility} We do not collect any new datasets as part of this paper. The goal of this work is to produce unlearning methods that can allow model providers to avoid having to retrain large language models from scratch. Thus we believe that the compute used for this project is reasonably justified. We also ran many of our pilot studies and ablations on the smaller Phi-1.5 model to reduce our energy and compute usage. 
\section*{Acknowledgments}
We gratefully acknowledge funding from the Natural Sciences and Engineering Research Council of Canada (NSERC) and the Canada CIFAR AI Chairs Program. Resources used in preparing this research were provided, in part, by the Province of Ontario, the Government of Canada through CIFAR, companies sponsoring the Vector Institute, Open Philanthropy, and through Schmidt Sciences via Roger Grosse's AI2050 Senior Fellowship and Nicolas Papernot's AI2050 Early Career Fellowship. Lev McKinney was supported in the form of a Constellation Visiting Fellowship, a NSERC Canada Graduate Scholarship-Master's, and by the Vector Scholarship in Artificial Intelligence, provided through the Vector Institute.  Anvith Thudi was supported by a Vanier Fellowship from NSERC. We would like to thank the following people for useful advice and discussion: Felix Dangel, Max Kaufmann, Zora Che, Stephen Casper, Andrew Wang, Stephen Zhao, Benson Li, Abhay Sheshadri, and Toryn Klassen.


\bibliographystyle{IEEEtran}
\bibliography{references}

@inproceedings{adam2019pytorch,
  author       = {Adam Paszke and
                  Sam Gross and
                  Francisco Massa and
                  Adam Lerer and
                  James Bradbury and
                  Gregory Chanan and
                  Trevor Killeen and
                  Zeming Lin and
                  Natalia Gimelshein and
                  Luca Antiga and
                  Alban Desmaison and
                  Andreas K{\"{o}}pf and
                  Edward Z. Yang and
                  Zachary DeVito and
                  Martin Raison and
                  Alykhan Tejani and
                  Sasank Chilamkurthy and
                  Benoit Steiner and
                  Lu Fang and
                  Junjie Bai and
                  Soumith Chintala},
  title        = {PyTorch: An Imperative Style, High-Performance Deep Learning Library},
  booktitle    = {Advances in Neural Information Processing Systems 32: Annual Conference
                  on Neural Information Processing Systems 2019, NeurIPS 2019},
  pages        = {8024--8035},
  year         = {2019},
  url          = {https://proceedings.neurips.cc/paper/2019/hash/bdbca288fee7f92f2bfa9f7012727740-Abstract.html},
}

@article{amari1998natural,
  author       = {Shun{-}Ichi Amari},
  title        = {Natural Gradient Works Efficiently in Learning},
  journal      = {Neural Computation},
  volume       = {10},
  number       = {2},
  pages        = {251--276},
  year         = {1998},
  publisher    = {{MIT} Press},
  doi          = {10.1162/089976698300017746}
}

@inproceedings{ba2017distributed,
  author       = {Lei Jimmy Ba and
                  Roger B. Grosse and
                  James Martens},
  title        = {Distributed Second-Order Optimization using Kronecker-Factored Approximations},
  booktitle    = {5th International Conference on Learning Representations, {ICLR} 2017,
                  Toulon, France, April 24-26, 2017, Conference Track Proceedings},
  publisher    = {OpenReview.net},
  year         = {2017},
  url          = {https://openreview.net/forum?id=SkkTMpjex}
}

@inproceedings{bae2022if,
  author       = {Juhan Bae and
                  Nathan Ng and
                  Alston Lo and
                  Marzyeh Ghassemi and
                  Roger B. Grosse},
  title        = {If Influence Functions are the Answer, Then What is the Question?},
  booktitle    = {Advances in Neural Information Processing Systems 35: Annual Conference
                  on Neural Information Processing Systems 2022, NeurIPS 2022},
  year         = {2022},
  url          = {https://openreview.net/forum?id=hzbguA9zMJ}
}

@inproceedings{bourtoule2021machine,
  author       = {Lucas Bourtoule and
                  Varun Chandrasekaran and
                  Christopher A. Choquette{-}Choo and
                  Hengrui Jia and
                  Adelin Travers and
                  Baiwu Zhang and
                  David Lie and
                  Nicolas Papernot},
  title        = {Machine Unlearning},
  booktitle    = {42nd {IEEE} Symposium on Security and Privacy, {SP} 2021, San Francisco,
                  CA, USA, May 24-27, 2021},
  pages        = {141--159},
  publisher    = {{IEEE}},
  year         = {2021},
  doi          = {10.1109/SP40001.2021.00019}
}

@inproceedings{carlini2021extracting,
  author       = {Nicholas Carlini and
                  Florian Tram{\`{e}}r and
                  Eric Wallace and
                  Matthew Jagielski and
                  Ariel Herbert{-}Voss and
                  Katherine Lee and
                  Adam Roberts and
                  Tom B. Brown and
                  Dawn Song and
                  {\'{U}}lfar Erlingsson and
                  Alina Oprea and
                  Colin Raffel},
  title        = {Extracting Training Data from Large Language Models},
  booktitle    = {30th {USENIX} Security Symposium, {USENIX} Security 2021},
  pages        = {2633--2650},
  publisher    = {{USENIX} Association},
  year         = {2021},
  url          = {https://www.usenix.org/conference/usenixsecurity21/presentation/carlini-extracting}
}

@inproceedings{carlini2024poisoning,
  author       = {Nicholas Carlini and
                  Matthew Jagielski and
                  Christopher A. Choquette{-}Choo and
                  Daniel Paleka and
                  Will Pearce and
                  Hyrum Anderson and
                  Andreas Terzis and
                  Kurt Thomas and
                  Florian Tram{\`{e}}r},
  title        = {Poisoning Web-Scale Training Datasets is Practical},
  booktitle    = {45th {IEEE} Symposium on Security and Privacy, {SP} 2024, San Francisco,
                  CA, USA, May 19-23, 2024},
  pages        = {407--425},
  publisher    = {{IEEE}},
  year         = {2024},
  doi          = {10.1109/SP54263.2024.00179}
}

@article{cooper2024machine,
  author       = {A. Feder Cooper and
                  Christopher A. Choquette{-}Choo and
                  Miranda Bogen and
                  Matthew Jagielski and
                  Katja Filippova and
                  Ken Ziyu Liu and
                  Alexandra Chouldechova and
                  Jamie Hayes and
                  Yangsibo Huang and
                  Niloofar Mireshghallah and
                  others},
  title        = {Machine Unlearning Doesn't Do What You Think: Lessons for Generative
                  {AI} Policy, Research, and Practice},
  journal      = {CoRR},
  volume       = {abs/2412.06966},
  year         = {2024},
  url          = {https://doi.org/10.48550/arXiv.2412.06966},
  doi          = {10.48550/arXiv.2412.06966},
  eprinttype   = {arXiv},
  eprint       = {2412.06966},
}

@misc{deeb2024unlearningmethodsremoveinformation,
      title={Do Unlearning Methods Remove Information from Language Model Weights?},
      author={Aghyad Deeb and Fabien Roger},
      year={2024},
      eprint={2410.08827},
      archivePrefix={arXiv},
      primaryClass={cs.LG},
      url={https://arxiv.org/abs/2410.08827},
}

@inproceedings{fan2024simplicity,
    title={Simplicity Prevails: Rethinking Negative Preference Optimization for {LLM} Unlearning},
    author={Chongyu Fan and Jiancheng Liu and Licong Lin and Jinghan Jia and Ruiqi Zhang and Song Mei and Sijia Liu},
    booktitle={Neurips Safe Generative AI Workshop 2024},
    year={2024},
    url={https://openreview.net/forum?id=pVACX02m0p}
}

@inproceedings{george2018fast,
  author       = {Thomas George and
                  C{\'{e}}sar Laurent and
                  Xavier Bouthillier and
                  Nicolas Ballas and
                  Pascal Vincent},
  title        = {Fast Approximate Natural Gradient Descent in a Kronecker Factored
                  Eigenbasis},
  booktitle    = {Advances in Neural Information Processing Systems 31: Annual Conference
                  on Neural Information Processing Systems 2018, NeurIPS 2018},
  pages        = {9573--9583},
  year         = {2018},
  url          = {https://proceedings.neurips.cc/paper/2018/hash/48000647b315f6f00f913caa757a70b3-Abstract.html}
}

@inproceedings{golatkar_eternal_2020,
  author       = {Aditya Golatkar and
                  Alessandro Achille and
                  Stefano Soatto},
  title        = {Eternal Sunshine of the Spotless Net: Selective Forgetting in Deep
                  Networks},
  booktitle    = {2020 {IEEE/CVF} Conference on Computer Vision and Pattern Recognition,
                  {CVPR} 2020},
  pages        = {9301--9309},
  publisher    = {{IEEE}},
  year         = {2020},
  doi          = {10.1109/CVPR42600.2020.00932},
  url          = {https://doi.org/10.1109/CVPR42600.2020.00932}
}

@inproceedings{guo2019certified,
  author       = {Chuan Guo and
                  Tom Goldstein and
                  Awni Y. Hannun and
                  Laurens van der Maaten},
  title        = {Certified Data Removal from Machine Learning Models},
  booktitle    = {Proceedings of the 37th International Conference on Machine Learning,
                  {ICML} 2020},
  series       = {Proceedings of Machine Learning Research},
  volume       = {119},
  pages        = {3832--3842},
  publisher    = {{PMLR}},
  year         = {2020},
  url          = {http://proceedings.mlr.press/v119/guo20c.html}
}

@inproceedings{hendrycks2021measuring,
title={Measuring Massive Multitask Language Understanding},
author={Dan Hendrycks and Collin Burns and Steven Basart and Andy Zou and Mantas Mazeika and Dawn Song and Jacob Steinhardt},
booktitle={International Conference on Learning Representations},
year={2021},
url={https://openreview.net/forum?id=d7KBjmI3GmQ}
}

@inproceedings{hessianfree2010,
  author       = {James Martens},
  title        = {Deep learning via Hessian-free optimization},
  booktitle    = {Proceedings of the 27th International Conference on Machine Learning,
                  {ICML} 2010},
  pages        = {735--742},
  publisher    = {Omnipress},
  year         = {2010},
  url          = {https://icml.cc/Conferences/2010/papers/458.pdf}
}

@inproceedings{huang-etal-2022-large,
    title = "Are Large Pre-Trained Language Models Leaking Your Personal Information?",
    author = "Huang, Jie  and
      Shao, Hanyin  and
      Chang, Kevin Chen-Chuan",
    editor = "Goldberg, Yoav  and
      Kozareva, Zornitsa  and
      Zhang, Yue",
    booktitle = "Findings of the Association for Computational Linguistics: EMNLP 2022",
    month = dec,
    year = "2022",
    address = "Abu Dhabi, United Arab Emirates",
    publisher = "Association for Computational Linguistics",
    url = "https://aclanthology.org/2022.findings-emnlp.148/",
    doi = "10.18653/v1/2022.findings-emnlp.148",
    pages = "2038--2047",
}

@inproceedings{ilharco2023taskarithmetic,
  author       = {Gabriel Ilharco and
                  Marco T{\'{u}}lio Ribeiro and
                  Mitchell Wortsman and
                  Suchin Gururangan and
                  Ludwig Schmidt and
                  Hannaneh Hajishirzi and
                  Ali Farhadi},
  title        = {Editing Models with Task Arithmetic},
  booktitle    = {The Eleventh International Conference on Learning Representations,
                  {ICLR} 2023, Kigali, Rwanda, May 1-5, 2023},
  publisher    = {OpenReview.net},
  year         = {2023},
  url          = {https://openreview.net/forum?id=6t0Kwf8-jrj}
}

@inproceedings{izzo2021approximate,
  author       = {Zachary Izzo and
                  Mary Anne Smart and
                  Kamalika Chaudhuri and
                  James Zou},
  title        = {Approximate Data Deletion from Machine Learning Models},
  booktitle    = {The 24th International Conference on Artificial Intelligence and Statistics,
                  {AISTATS} 2021, April 13-15, 2021, Virtual Event},
  series       = {Proceedings of Machine Learning Research},
  volume       = {130},
  pages        = {2008--2016},
  publisher    = {{PMLR}},
  year         = {2021},
  url          = {http://proceedings.mlr.press/v130/izzo21a.html}
}

@inproceedings{jang2022knowledge,
  author       = {Joel Jang and
                  Dongkeun Yoon and
                  Sohee Yang and
                  Sungmin Cha and
                  Moontae Lee and
                  Lajanugen Logeswaran and
                  Minjoon Seo},
  title        = {Knowledge Unlearning for Mitigating Privacy Risks in Language Models},
  booktitle    = {Proceedings of the 61st Annual Meeting of the Association for Computational Linguistics (Volume 1: Long Papers), {ACL} 2023},
  pages        = {14389--14408},
  publisher    = {Association for Computational Linguistics},
  year         = {2023},
  url          = {https://doi.org/10.18653/v1/2023.acl-long.805},
  doi          = {10.18653/v1/2023.acl-long.805},
}

@inproceedings{hsu2024safelora,
  author       = {Chia{-}Yi Hsu and
                  Yu{-}Lin Tsai and
                  Chih{-}Hsun Lin and
                  Pin{-}Yu Chen and
                  Chia{-}Mu Yu and
                  Chun{-}Ying Huang},
  editor       = {Amir Globersons and
                  Lester Mackey and
                  Danielle Belgrave and
                  Angela Fan and
                  Ulrich Paquet and
                  Jakub M. Tomczak and
                  Cheng Zhang},
  title        = {Safe LoRA: The Silver Lining of Reducing Safety Risks when Finetuning
                  Large Language Models},
  booktitle    = {Advances in Neural Information Processing Systems 38: Annual Conference
                  on Neural Information Processing Systems 2024, NeurIPS 2024, Vancouver,
                  BC, Canada, December 10 - 15, 2024},
  year         = {2024},
  url          = {http://papers.nips.cc/paper\_files/paper/2024/hash/77baa7c2a3a675823e89131698fd6e19-Abstract-Conference.html},
  bibsource    = {dblp computer science bibliography, https://dblp.org}
}

@inproceedings{haung2025antidote,
  author       = {Tiansheng Huang and
                  Gautam Bhattacharya and
                  Pratik Joshi and
                  Joshua Kimball and
                  Ling Liu},
  editor       = {Aarti Singh and
                  Maryam Fazel and
                  Daniel Hsu and
                  Simon Lacoste{-}Julien and
                  Felix Berkenkamp and
                  Tegan Maharaj and
                  Kiri Wagstaff and
                  Jerry Zhu},
  title        = {Antidote: Post-fine-tuning Safety Alignment for Large Language Models
                  against Harmful Fine-tuning Attack},
  booktitle    = {Forty-second International Conference on Machine Learning, {ICML}
                  2025, Vancouver, BC, Canada, July 13-19, 2025},
  series       = {Proceedings of Machine Learning Research},
  volume       = {267},
  publisher    = {{PMLR} / OpenReview.net},
  year         = {2025},
  url          = {https://proceedings.mlr.press/v267/huang25b.html},
  bibsource    = {dblp computer science bibliography, https://dblp.org}
}

@inproceedings{jia2023model,
  author       = {Jinghan Jia and
                  Jiancheng Liu and
                  Parikshit Ram and
                  Yuguang Yao and
                  Gaowen Liu and
                  Yang Liu and
                  Pranay Sharma and
                  Sijia Liu},
  title        = {Model Sparsity Can Simplify Machine Unlearning},
  booktitle    = {Advances in Neural Information Processing Systems 36: Annual Conference
                  on Neural Information Processing Systems 2023, NeurIPS 2023, New Orleans,
                  LA, USA, December 10-16, 2023},
  year         = {2023},
  url          = {https://openreview.net/forum?id=5FedOMqMZu}
}

@article{jiang2023mistral7b,
  author       = {Albert Q. Jiang and
                  Alexandre Sablayrolles and
                  Arthur Mensch and
                  Chris Bamford and
                  Devendra Singh Chaplot and
                  Diego de las Casas and
                  Florian Bressand and
                  Gianna Lengyel and
                  Guillaume Lample and
                  Lucile Saulnier and
                  L{\'{e}}lio Renard Lavaud and
                  Marie{-}Anne Lachaux and
                  Pierre Stock and
                  Teven Le Scao and
                  Thibaut Lavril and
                  Thomas Wang and
                  Timoth{\'{e}}e Lacroix and
                  William El Sayed},
  title        = {Mistral 7B},
  journal      = {CoRR},
  volume       = {abs/2310.06825},
  year         = {2023},
  url          = {https://doi.org/10.48550/arXiv.2310.06825},
  doi          = {10.48550/arXiv.2310.06825},
  eprinttype   = {arXiv},
  eprint       = {2310.06825},
}

@inproceedings{kingma-ba-adam2015,
  author       = {Diederik P. Kingma and
                  Jimmy Ba},
  title        = {Adam: {A} Method for Stochastic Optimization},
  booktitle    = {Proceedings of the 3rd International Conference on Learning Representations, {ICLR}},
  year         = {2015},
  url          = {http://arxiv.org/abs/1412.6980}
}

@inproceedings{koh_understanding_2020,
  author       = {Pang Wei Koh and
                  Percy Liang},
  title        = {Understanding Black-box Predictions via Influence Functions},
  booktitle    = {Proceedings of the 34th International Conference on Machine Learning,
                  {ICML} 2017},
  series       = {Proceedings of Machine Learning Research},
  volume       = {70},
  pages        = {1885--1894},
  publisher    = {{PMLR}},
  year         = {2017},
  url          = {http://proceedings.mlr.press/v70/koh17a.html}
}

@inproceedings{kurmanji2023towards,
  author       = {Meghdad Kurmanji and
                  Peter Triantafillou and
                  Jamie Hayes and
                  Eleni Triantafillou},
  title        = {Towards Unbounded Machine Unlearning},
  booktitle    = {Advances in Neural Information Processing Systems 36: Annual Conference
                  on Neural Information Processing Systems 2023, NeurIPS 2023, New Orleans,
                  LA, USA, December 10-16, 2023},
  year         = {2023},
  url          = {https://openreview.net/forum?id=OveBaTtUAT}
}

@inproceedings{lin2004rouge,
  author       = {Chin{-}Yew Lin},
  title        = {{ROUGE:} {A} Package for Automatic Evaluation of Summaries},
  booktitle    = {Text Summarization Branches Out, Proceedings of the {ACL-04} Workshop,
                  Barcelona, Spain, July 25-26, 2004},
  pages        = {74--81},
  publisher    = {Association for Computational Linguistics},
  year         = {2004},
  url          = {https://aclanthology.org/W04-1013/}
}

@inproceedings{liu2022continual,
  author       = {Bo Liu and
                  Qiang Liu and
                  Peter Stone},
  title        = {Continual Learning and Private Unlearning},
  booktitle    = {Conference on Lifelong Learning Agents, CoLLAs 2022, 22-24 August 2022,
                  Montreal, QC, Canada},
  series       = {Proceedings of Machine Learning Research},
  volume       = {199},
  pages        = {243--254},
  publisher    = {{PMLR}},
  year         = {2022},
  url          = {https://proceedings.mlr.press/v199/liu22a.html}
}

@inproceedings{liu2024sophiascalablestochasticsecondorder,
  author       = {Hong Liu and
                  Zhiyuan Li and
                  David Leo Wright Hall and
                  Percy Liang and
                  Tengyu Ma},
  title        = {Sophia: {A} Scalable Stochastic Second-order Optimizer for Language
                  Model Pre-training},
  booktitle    = {The Twelfth International Conference on Learning Representations,
                  {ICLR} 2024, Vienna, Austria, May 7-11, 2024},
  publisher    = {OpenReview.net},
  year         = {2024},
  url          = {https://openreview.net/forum?id=3HLwzzdQxm}
}

@article{martensNewInsightsPerspectives2020,
  author       = {James Martens},
  title        = {New Insights and Perspectives on the Natural Gradient Method},
  journal      = {Journal of Machine Learning Research},
  volume       = {21},
  pages        = {146:1--146:76},
  year         = {2020},
  url          = {http://jmlr.org/papers/v21/17-678.html}
}

@inproceedings{martensOptimizingNeuralNetworks2015,
  author       = {James Martens and
                  Roger B. Grosse},
  title        = {Optimizing Neural Networks with Kronecker-factored Approximate Curvature},
  booktitle    = {Proceedings of the 32nd International Conference on Machine Learning,
                  {ICML} 2015},
  series       = {{JMLR} Workshop and Conference Proceedings},
  volume       = {37},
  pages        = {2408--2417},
  publisher    = {JMLR.org},
  year         = {2015},
  url          = {http://proceedings.mlr.press/v37/martens15.html}
}

@inproceedings{park2023trak,
  author       = {Sung Min Park and
                  Kristian Georgiev and
                  Andrew Ilyas and
                  Guillaume Leclerc and
                  Aleksander Madry},
  title        = {{TRAK:} Attributing Model Behavior at Scale},
  booktitle    = {Proceedings of the 40th International Conference on Machine Learning,
                  {ICML} 2023, Honolulu, Hawaii, USA, July 23-29, 2023},
  series       = {Proceedings of Machine Learning Research},
  volume       = {202},
  pages        = {27074--27113},
  publisher    = {{PMLR}},
  year         = {2023},
  url          = {https://proceedings.mlr.press/v202/park23c.html}
}

@misc{qwen2025qwen25technicalreport,
  author       = {{Qwen Team} and
                  An Yang and
                  Baosong Yang and
                  Beichen Zhang and
                  Binyuan Hui and
                  Bo Zheng and
                  Bowen Yu and
                  Chengyuan Li and
                  Dayiheng Liu and
                  Fei Huang and
                  Haoran Wei and
                  Huan Lin and
                  Jian Yang and
                  Jianhong Tu and
                  Jianwei Zhang and
                  Jianxin Yang and
                  Jiaxi Yang and
                  Jingren Zhou and
                  Junyang Lin and
                  others},
  title        = {Qwen2.5 Technical Report},
  year         = {2024},
  eprint       = {2412.15115},
  archivePrefix= {arXiv},
  primaryClass = {cs.CL},
  url          = {https://arxiv.org/abs/2412.15115}
}

@inproceedings{rosati2024representation,
    title={Representation Noising: A Defence Mechanism Against Harmful Finetuning},
    author={Domenic Rosati and Jan Wehner and Kai Williams and Lukasz Bartoszcze and Robie Gonzales and carsten maple and Subhabrata Majumdar and Hassan Sajjad and Frank Rudzicz},
    booktitle={The Thirty-eighth Annual Conference on Neural Information Processing Systems},
    year={2024},
    url={https://openreview.net/forum?id=eP9auEJqFg}
}

@inproceedings{singh2020woodfisher,
  author       = {Sidak Pal Singh and
                  Dan Alistarh},
  title        = {WoodFisher: Efficient Second-Order Approximation for Neural Network
                  Compression},
  booktitle    = {Advances in Neural Information Processing Systems 33: Annual Conference
                  on Neural Information Processing Systems 2020, NeurIPS 2020, December
                  6-12, 2020, virtual},
  year         = {2020},
  url          = {https://proceedings.neurips.cc/paper/2020/hash/d1ff1ec86b62cd5f3903ff19c3a326b2-Abstract.html}
}

@article{textbooks2,
  author       = {Yuanzhi Li and
                  S{\'{e}}bastien Bubeck and
                  Ronen Eldan and
                  Allie Del Giorno and
                  Suriya Gunasekar and
                  Yin Tat Lee},
  title        = {Textbooks Are All You Need {II:} phi-1.5 technical report},
  journal      = {CoRR},
  volume       = {abs/2309.05463},
  year         = {2023},
  url          = {https://doi.org/10.48550/arXiv.2309.05463},
  doi          = {10.48550/arXiv.2309.05463},
  eprinttype   = {arXiv},
  eprint       = {2309.05463},
}

@inproceedings{thudi2022necessity,
  author       = {Anvith Thudi and
                  Hengrui Jia and
                  Ilia Shumailov and
                  Nicolas Papernot},
  title        = {On the Necessity of Auditable Algorithmic Definitions for Machine
                  Unlearning},
  booktitle    = {31st {USENIX} Security Symposium, {USENIX} Security 2022, Boston, MA,
                  USA, August 10-12, 2022},
  pages        = {4007--4022},
  publisher    = {{USENIX} Association},
  year         = {2022},
  url          = {https://www.usenix.org/conference/usenixsecurity22/presentation/thudi}
}

@inproceedings{thudi2024gradients,
  author       = {Anvith Thudi and
                  Hengrui Jia and
                  Casey Meehan and
                  Ilia Shumailov and
                  Nicolas Papernot},
  title        = {Gradients Look Alike: Sensitivity is Often Overestimated in {DP-SGD}},
  booktitle    = {33rd {USENIX} Security Symposium, {USENIX} Security 2024, Philadelphia,
                  PA, USA, August 14-16, 2024},
  pages        = {973--990},
  publisher    = {{USENIX} Association},
  year         = {2024},
  url          = {https://www.usenix.org/conference/usenixsecurity24/presentation/thudi}
}

@article{touvron2023llama2openfoundation,
  author       = {Hugo Touvron and
                  Louis Martin and
                  Kevin Stone and
                  Peter Albert and
                  Amjad Almahairi and
                  Yasmine Babaei and
                  Nikolay Bashlykov and
                  Soumya Batra and
                  Prajjwal Bhargava and
                  Shruti Bhosale and
                  others},
  title        = {Llama 2: Open Foundation and Fine-Tuned Chat Models},
  journal      = {CoRR},
  volume       = {abs/2307.09288},
  year         = {2023},
  url          = {https://doi.org/10.48550/arXiv.2307.09288},
  doi          = {10.48550/arXiv.2307.09288},
  eprinttype   = {arXiv},
  eprint       = {2307.09288},
}

@inproceedings{wikitext2017,
    title={Pointer Sentinel Mixture Models},
    author={Stephen Merity and Caiming Xiong and James Bradbury and Richard Socher},
    booktitle={International Conference on Learning Representations},
    year={2017},
    url={https://openreview.net/forum?id=Byj72udxe}
}

@article{zhao2023pytorch,
  author       = {Yanli Zhao and
                  Andrew Gu and
                  Rohan Varma and
                  Liang Luo and
                  Chien{-}Chin Huang and
                  Min Xu and
                  Less Wright and
                  Hamid Shojanazeri and
                  Myle Ott and
                  Sam Shleifer and
                  Alban Desmaison and
                  Can Balioglu and
                  Pritam Damania and
                  Bernard Nguyen and
                  Geeta Chauhan and
                  Yuchen Hao and
                  Ajit Mathews and
                  Shen Li},
  title        = {PyTorch {FSDP:} Experiences on Scaling Fully Sharded Data Parallel},
  journal      = {Proc. {VLDB} Endow.},
  volume       = {16},
  number       = {12},
  pages        = {3848--3860},
  year         = {2023},
  doi          = {10.14778/3611540.3611569}
}

@article{zheng2023judging,
  title={Judging LLM-as-a-Judge with MT-Bench and Chatbot Arena},
  author={Zheng, Lianmin and Chiang, Wei-Lin and Sheng, Ying and Zhuang, Siyuan and Wu, Zhanghao and Zhuang, Yonghao and Lin, Zi and Li, Zhuohan and Li13, Dacheng and Xing35, Eric P and others},
  journal={arXiv preprint arXiv:2306.05685},
  year={2023}
}

@inproceedings{jia_soul_2024,
  author       = {Jinghan Jia and
                  Yihua Zhang and
                  Yimeng Zhang and
                  Jiancheng Liu and
                  Bharat Runwal and
                  James Diffenderfer and
                  Bhavya Kailkhura and
                  Sijia Liu},
  title        = {{SOUL:} Unlocking the Power of Second-Order Optimization for {LLM}
                  Unlearning},
  booktitle    = {Proceedings of the 2024 Conference on Empirical Methods in Natural
                  Language Processing, {EMNLP} 2024, Miami, FL, USA, November 12-16,
                  2024},
  pages        = {4276--4292},
  publisher    = {Association for Computational Linguistics},
  year         = {2024},
  url          = {https://aclanthology.org/2024.emnlp-main.245/},
  doi          = {10.18653/v1/2024.emnlp-main.245}
}

@inproceedings{li_wmdp_2024,
  author       = {Nathaniel Li and
                  Alexander Pan and
                  Anjali Gopal and
                  Summer Yue and
                  Daniel Berrios and
                  Alice Gatti and
                  Justin D. Li and
                  Ann{-}Kathrin Dombrowski and
                  Shashwat Goel and
                  Long Phan and
                  Gabriel Mukobi and
                  Nathan Helm{-}Burger and
                  Rassin Lababidi and
                  Lennart Justen and
                  Andrew B. Liu and
                  Michael Chen and
                  Isabelle Barrass and
                  Oliver Zhang and
                  Xiaoyuan Zhu and
                  Rishub Tamirisa and
                  Bhrugu Bharathi and
                  Adam Khoja and
                  Zhenqi Zhao and
                  Ariel Herbert{-}Voss and
                  Cort B. Breuer and
                  Samuel Marks and
                  Oam Patel and
                  Andy Zou and
                  Mantas Mazeika and
                  Zifan Wang and
                  Palash Oswal and
                  Weiran Lin and
                  Adam A. Hunt and
                  Justin Tienken{-}Harder and
                  Kevin Y. Shih and
                  Kemper Talley and
                  John Guan and
                  Russell Kaplan and
                  Ian Steneker and
                  David Campbell and
                  Brad Jokubaitis and
                  Alex Levinson and
                  Jean Wang and
                  William Qian and
                  Kallol Krishna Karmakar and
                  Steven Basart and
                  Stephen Fitz and
                  Mindy Levine and
                  Ponnurangam Kumaraguru and
                  Uday Tupakula and
                  Vijay Varadharajan and
                  Ruoyu Wang and
                  Yan Shoshitaishvili and
                  Jimmy Ba and
                  Kevin M. Esvelt and
                  Alexandr Wang and
                  Dan Hendrycks},
  title        = {The {WMDP} Benchmark: Measuring and Reducing Malicious Use With Unlearning},
  booktitle    = {Proceedings of the 41st International Conference on Machine Learning,
                  {ICML} 2024, Vienna, Austria, July 21-27, 2024},
  series       = {Proceedings of Machine Learning Research},
  publisher    = {{PMLR}},
  year         = {2024},
  url          = {https://openreview.net/forum?id=sX13gVbFCC}
}

@article{liu_rethinking_2024,
  author       = {Sijia Liu and
                  Yuanshun Yao and
                  Jinghan Jia and
                  Stephen Casper and
                  Nathalie Baracaldo and
                  Peter Hase and
                  Yuguang Yao and
                  Chris Yuhao Liu and
                  Xiaojun Xu and
                  Hang Li and
                  Kush R. Varshney and
                  Mohit Bansal and
                  Sanmi Koyejo and
                  Yang Liu},
  title        = {Rethinking machine unlearning for large language models},
  journal      = {Nature Machine Intelligence},
  volume       = {7},
  number       = {2},
  pages        = {181--194},
  year         = {2025},
  doi          = {10.1038/s42256-025-00985-0}
}

@inproceedings{qi2024evaluating,
  author       = {Xiangyu Qi and
                  Boyi Wei and
                  Nicholas Carlini and
                  Yangsibo Huang and
                  Tinghao Xie and
                  Luxi He and
                  Matthew Jagielski and
                  Milad Nasr and
                  Prateek Mittal and
                  Peter Henderson},
  title        = {On Evaluating the Durability of Safeguards for Open-Weight {LLMs}},
  booktitle    = {The Thirteenth International Conference on Learning Representations,
                  {ICLR} 2025},
  publisher    = {OpenReview.net},
  year         = {2025},
  url          = {https://openreview.net/forum?id=fXJCqdUSVG}
}

@inproceedings{shi_muse_2024,
  author       = {Weijia Shi and
                  Jaechan Lee and
                  Yangsibo Huang and
                  Sadhika Malladi and
                  Jieyu Zhao and
                  Ari Holtzman and
                  Daogao Liu and
                  Luke Zettlemoyer and
                  Noah A. Smith and
                  Chiyuan Zhang},
  title        = {{MUSE:} Machine Unlearning Six-Way Evaluation for Language Models},
  booktitle    = {The Thirteenth International Conference on Learning Representations,
                  {ICLR} 2025},
  publisher    = {OpenReview.net},
  year         = {2025},
  url          = {https://openreview.net/forum?id=O2JxWkJByy}
}

@inproceedings{tamirisa2024tamperresistantsafeguardsopenweightllms,
  author       = {Rishub Tamirisa and
                  Bhrugu Bharathi and
                  Long Phan and
                  Andy Zhou and
                  Alice Gatti and
                  Tarun Suresh and
                  Maxwell Lin and
                  Justin Wang and
                  Rowan Wang and
                  Ron Arel and
                  Andy Zou and
                  Dawn Song and
                  Bo Li and
                  Dan Hendrycks and
                  Mantas Mazeika},
  title        = {Tamper-Resistant Safeguards for Open-Weight {LLMs}},
  booktitle    = {The Thirteenth International Conference on Learning Representations,
                  {ICLR} 2025},
  publisher    = {OpenReview.net},
  year         = {2025},
  url          = {https://openreview.net/forum?id=4FIjRodbW6}
}

@inproceedings{yao_large_2024,
  author       = {Yuanshun Yao and
                  Xiaojun Xu and
                  Yang Liu},
  title        = {Large Language Model Unlearning},
  booktitle    = {Advances in Neural Information Processing Systems 37: Annual Conference
                  on Neural Information Processing Systems 2024, NeurIPS 2024},
  year         = {2024},
  url          = {https://neurips.cc/virtual/2024/poster/93184}
}

@inproceedings{yao_machine_2024,
  author       = {Jin Yao and
                  Eli Chien and
                  Minxin Du and
                  Xinyao Niu and
                  Tianhao Wang and
                  Zezhou Cheng and
                  Xiang Yue},
  title        = {Machine Unlearning of Pre-trained Large Language Models},
  booktitle    = {Proceedings of the 62nd Annual Meeting of the Association for
                  Computational Linguistics (Volume 1: Long Papers), {ACL} 2024,
                  Bangkok, Thailand, August 11-16, 2024},
  pages        = {8403--8419},
  publisher    = {Association for Computational Linguistics},
  year         = {2024},
  url          = {https://aclanthology.org/2024.acl-long.457/},
  doi          = {10.18653/v1/2024.acl-long.457}
}

@misc{alpaca,
  author = {Rohan Taori and Ishaan Gulrajani and Tianyi Zhang and Yann Dubois and Xuechen Li and Carlos Guestrin and Percy Liang and Tatsunori B. Hashimoto},
  title = {Stanford Alpaca: An Instruction-following LLaMA model},
  year = {2023},
  publisher = {GitHub},
  journal = {GitHub repository},
  howpublished = {\url{https://github.com/tatsu-lab/stanford_alpaca}},
}

@software{bae2024kronfluence,
  author       = {Juhan Bae},
  title        = {Kronfluence: Influence Functions with Kronecker-factored Approximate
                  Curvature},
  year         = {2024},
  publisher    = {GitHub},
  url          = {https://github.com/pomonam/kronfluence},
  doi          = {10.5281/zenodo.13131049}
}

@misc{dangel2025positioncurvaturematricesdemocratized,
  author       = {Felix Dangel and
                  Runa Eschenhagen and
                  Weronika Ormaniec and
                  Andres Fernandez and
                  Lukas Tatzel and
                  Agustinus Kristiadi},
  title        = {Position: Curvature Matrices Should Be Democratized via Linear Operators},
  year         = {2025},
  eprint       = {2501.19183},
  archivePrefix= {arXiv},
  primaryClass = {cs.LG},
  url          = {https://arxiv.org/abs/2501.19183}
}

@article{elm2024,
  publtype={informal},
  author={Rohit Gandikota and Sheridan Feucht and Samuel Marks and David Bau},
  title={Erasing Conceptual Knowledge from Language Models},
  year={2024},
  cdate={1704067200000},
  journal={CoRR},
  volume={abs/2410.02760},
  url={https://doi.org/10.48550/arXiv.2410.02760}
}

@article{georgiev2024attribute,
  author       = {Kristian Georgiev and
                  Roy Rinberg and
                  Sung Min Park and
                  Shivam Garg and
                  Andrew Ilyas and
                  Aleksander Madry and
                  Seth Neel},
  title        = {Attribute-to-Delete: Machine Unlearning via Datamodel Matching},
  journal      = {CoRR},
  volume       = {abs/2410.23232},
  year         = {2024},
  eprint       = {2410.23232},
  archivePrefix= {arXiv},
  primaryClass = {cs.LG},
  url          = {https://arxiv.org/abs/2410.23232}
}

@misc{grosse2022metrics,
    author = {Grosse, Roger},
    title = {Metrics},
    year = {2022},
    note = {Chapter 3 in CSC2541: Topics in Machine Learning: Neural Net Training Dynamics, Course Notes},
    institution = {University of Toronto},
    type = {Course Notes},
    howpublished = {\url{https://www.cs.toronto.edu/~rgrosse/courses/csc2541_2022/}},
}

@misc{grosse2023studyinglargelanguagemodel,
  author       = {Roger Grosse and
                  Juhan Bae and
                  Cem Anil and
                  Nelson Elhage and
                  Alex Tamkin and
                  Amirhossein Tajdini and
                  Benoit Steiner and
                  Dustin Li and
                  Esin Durmus and
                  Ethan Perez and
                  Evan Hubinger and
                  Kamil\.{e} Luko\v{s}i\={u}t\.{e} and
                  Karina Nguyen and
                  Nicholas Joseph and
                  Sam McCandlish and
                  Jared Kaplan and
                  Samuel R. Bowman},
  title        = {Studying Large Language Model Generalization with Influence Functions},
  year         = {2023},
  eprint       = {2308.03296},
  archiveprefix = {arXiv},
  primaryclass = {cs.LG},
  url          = {https://arxiv.org/abs/2308.03296}
}

@misc{gu_second-order_2024,
	title = {Second-{Order} {Information} {Matters}: {Revisiting} {Machine} {Unlearning} for {Large} {Language} {Models}},
	shorttitle = {Second-{Order} {Information} {Matters}},
	url = {http://arxiv.org/abs/2403.10557},
	publisher = {arXiv},
	author = {Gu, Kang and Rashid, Md Rafi Ur and Sultana, Najrin and Mehnaz, Shagufta},
	month = mar,
	year = {2024},
	note = {arXiv:2403.10557 [cs]},
}

@misc{maini_tofu_2024,
	title = {{TOFU}: {A} {Task} of {Fictitious} {Unlearning} for {LLMs}},
	shorttitle = {{TOFU}},
	url = {http://arxiv.org/abs/2401.06121},
	doi = {10.48550/arXiv.2401.06121},
	publisher = {arXiv},
	author = {Maini, Pratyush and Feng, Zhili and Schwarzschild, Avi and Lipton, Zachary C. and Kolter, J. Zico},
	month = jan,
	year = {2024},
	note = {arXiv:2401.06121 [cs]},
}

@article{shazeer2020glu,
  author       = {Noam Shazeer},
  title        = {{GLU} Variants Improve Transformer},
  journal      = {CoRR},
  volume       = {abs/2002.05202},
  year         = {2020},
  eprint       = {2002.05202},
  archiveprefix = {arXiv},
  url          = {https://arxiv.org/abs/2002.05202}
}

@misc{souly2025poisoningattacksllmsrequire,
  author       = {Alexandra Souly and
                  Javier Rando and
                  Ed Chapman and
                  Xander Davies and
                  Burak Hasircioglu and
                  Ezzeldin Shereen and
                  Carlos Mougan and
                  Vasilios Mavroudis and
                  Erik Jones and
                  Chris Hicks and
                  Nicholas Carlini and
                  Yarin Gal and
                  Robert Kirk},
  title        = {Poisoning Attacks on {LLMs} Require a Near-constant Number of Poison
                  Samples},
  year         = {2025},
  eprint       = {2510.07192},
  archivePrefix= {arXiv},
  primaryClass = {cs.LG},
  url          = {https://arxiv.org/abs/2510.07192}
}

@misc{tunstall2023zephyrdirectdistillationlm,
  author       = {Lewis Tunstall and
                  Edward Beeching and
                  Nathan Lambert and
                  Nazneen Rajani and
                  Kashif Rasul and
                  Younes Belkada and
                  Shengyi Huang and
                  Leandro von Werra and
                  Cl{\'{e}}mentine Fourrier and
                  Nathan Habib and
                  Nathan Sarrazin and
                  Omar Sanseviero and
                  Alexander M. Rush and
                  Thomas Wolf},
  title        = {Zephyr: Direct Distillation of {LM} Alignment},
  year         = {2023},
  eprint       = {2310.16944},
  archiveprefix = {arXiv},
  primaryclass = {cs.LG},
  url          = {https://arxiv.org/abs/2310.16944}
}

@misc{zhang2024npo,
      title={Negative Preference Optimization: From Catastrophic Collapse to Effective Unlearning},
      author={Ruiqi Zhang and Licong Lin and Yu Bai and Song Mei},
      year={2024},
      eprint={2404.05868},
      archivePrefix={arXiv},
      primaryClass={cs.LG},
      url={https://arxiv.org/abs/2404.05868},
}
\newpage
\onecolumn
\appendix
\begin{table*}[h]
\centering
\caption{Notation}
\label{tab:notation}
\begin{tabular}{cl}
\toprule
\textbf{Symbol} & \textbf{Description} \\
\midrule
\multicolumn{2}{l}{\textit{Data}} \\
$D$ & The full training dataset used to train the model \\
$D_F$ & The forget set of examples which is representative of a broader distribution we want to suppress.\\
$D_R$ & The retain set of examples which is representative of a broader distribution we want to maintain the same output behavior on.\\
$x$ & An input to out network e.g. a sequence of token ids. \\
$y$ & The labels of a training example. \\

\multicolumn{2}{l}{\textit{Model and Parameters}} \\
$\theta$ & The parameters of our model. \\
$\theta^*$ & The optimal parameters of our model on the training objective. \\
$\theta_D$ & The parameters of a model trained on the entire training set using the training process $\mathcal{T}$. \\
$\theta_{D_R}$ & The parameters of a model trained on the retain set using the training process $\mathcal{T}$.\\
$\delta \theta$ & A small perturbation on our weights.\\
$f(x; \theta)$ & A function representing our LLM. It takes as arguments our datum $x$ and parameters $\theta$, and returns the models logits $z$.\\
$z$ & The language models logits i.e. the unnormalized log probabilities over next tokens.\\
$W$ & The weights of a linear layer inside a neural network in homogeneous coordinates.\\
$s$ & The activations of the network immediately \textit{after} the linear layer.\\
$a$ & The activations of the network immediately \textit{before} the linear layer.\\

\multicolumn{2}{l}{\textit{Loss and Objectives}} \\
$\mathcal{L}(\theta)$ & The training objective used to train a model. Typically the cross entropy loss. \\
$\mathcal{L}(\epsilon, \theta)$ & The training objective used to train a model. With the forget set $D_F$ reweighed by $\epsilon$, e.g. $\mathcal{L}(1, \theta) = \mathcal{L}( \theta)$. \\
$\mathcal{L}_F(\theta)$ & The forget objective we are maximizing for output suppression.\\
$\ell(z, y)$ & The per token negative log-likelihood $-z_y + \log(\sum_i e^{z_i})$.\\
$\ell^{(margin)}(z, y)$ & The per token margin loss $-z_y + \log(\sum_{i\neq y} e^{z_i})$.\\

\multicolumn{2}{l}{\textit{Probability and Distributions}} \\
$\rho (.|z)$ & A categorical distribution parameterized by the unnormalized log probabilities $z$ i.e. $\rho (y|z) = \frac{e^{z_y}}{\sum_i e^{z_i}}$ \\
$KL(p,q)$ & the KL divergence between distribution p and q.\\
$P_\theta(u|x)$ & The probability of assigned by auto-regressive language model to a sequence of tokens $u$ with a prompt $x$. \\
\multicolumn{2}{l}{\textit{Gradients and Derivatives}} \\
$\nabla_\theta$ & The gradient operator with respect to the parameters $\theta$.\\
$\bar \nabla_\theta$ & The natural gradient operator with respect to the parameters $\theta$ (see Sec \ref{sec:natgrad}\\
$\pgrad{c}(x; \theta)$ & Is the derivative with respect to an variable $c$ in the computation graph of $\log(\rho(\hat y| f(x;\theta))$ where $\hat{y} \sim \rho(\hat y| f(x;\theta))$. \\ 
& We will sometimes drop the $(x; \theta)$ when implied by the context. \\
\multicolumn{2}{l}{\textit{Hessian and Curvature Matrices}} \\
$H_\theta$ & The Hessian of the objective function on the dataset with respect to the model parameters $\theta$. \\
$G_\theta$ & The Hessian of the objective function on the dataset with respect to the model parameters $\theta$.\\

$H_z$ & The Hessian of the loss function with respect to the logits $z$.\\
$F_z$ & The Fisher information matrix of a probability distribution $\rho(.|z)$. \\

\multicolumn{2}{l}{\textit{K-FAC/EK-FAC Factors}} \\
$\mathbf{A}$ & Unentered covariance of the activations $a$ from just before the linear layer.\\
$\mathbf{S}$ & Unentered covariance of the post activation pseudo gradients $\mathcal D s$.\\
$(Q_M, \Lambda_M)$ & The spectral decomposition of the symmetric matrix $M$. \\
$\Lambda$ & A diagonal matrix containing the Gauss-Newton Hessian's approximate eigenvalues. \\
\multicolumn{2}{l}{\textit{Functions and Operators}} \\
$\mathcal{T}(D')$ & A deterministic training process which takes a dataset $D'$ as a function and returns and optimal set of model weights $\theta^*$.\\
$r(\epsilon)$ & the change in our optimal model parameters after up reweighting the forget set by $\epsilon$. \\
$\hat{r}(\epsilon)_H$ & An approximation to the change in model parameters using the models Hessian matrix $H$ see Sec \ref{sec:prelim}. \\
$\otimes$ & The kronecker product.\\
$\textrm{vec}(M)$ & The result of taking each of the matrix $M \in \mathbb{R}^{m\times n}$'s columns and stacking them to form a column vector $\textrm{vec}(M) \in \mathbb{R}^{(mn)}$. \\
\bottomrule
\end{tabular}
\end{table*}

\newpage
\section*{Hessian benchmarking across model scales}
\label{sec:hessian_benchmarking}

\begin{figure*}[h]
    \centering
    \includegraphics[width=\linewidth]{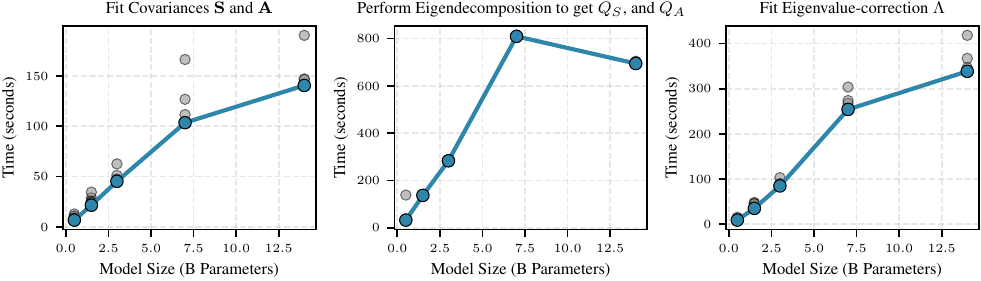}
    \caption{\textbf{The time required to fit the EK-FAC Hessian scales close to linearly with model size.} Here we show how three stages of fitting the EK-FAC Hessian approximation scale with model size; see Section~\ref{sec:prelim} for a breakdown of how each of these stages is implemented. The gray dots show sub-optimal configurations. Note that fitting the covariances (right) and the eigendecomposition (middle) are required in both K-FAC and EK-FAC, while the eigenvalue correction is only used in EK-FAC.}
    \label{fig:benchmarking}
\end{figure*}

To build on our analysis of the time complexity of EK-FAC and K-FAC in Section~\ref{sec:scaling_analysis}, we run additional empirical experiments showing how the time required to fit our Gauss-Newton Hessian approximation scales with model size and that the method scales to medium-size open-weight models on readily available hardware.

To accomplish this we make use of an adapted version of the kronfluence~\cite{bae2024kronfluence} library which provides a more scalable implementation of EK-FAC than CurvLinOps~\cite{dangel2025positioncurvaturematricesdemocratized}. This improved implementation shards the covariance factors $\mathbf{A}, \mathbf{S}$, their eigenvectors $Q_A$ and $Q_S$, as well as the approximate eigenvalues $\Lambda$ for each linear layer across all GPUs. They are only gathered when immediately needed for computations. This implementation essentially applies the fully sharded data parallel (FSDP) strategy to the various factors involved in Hessian computation, we also use FSDP~\cite{zhao2023pytorch} to shard our model weights. We intend to make a version of kronfluence implementing K-FADE and this Hessian sharding strategy. \footnote{\href{https://github.com/levmckinney/kronfluence}{https://github.com/levmckinney/kronfluence}}

Our experiments are performed on the Qwen 2.5~\cite{qwen2025qwen25technicalreport} scaling suite: 0.5B, 1.5B, 3B, 7B, and 14B models. For hardware we use an 8xH100 node with SXM5 connectivity. We fit our Hessians on 1000 sequences of length 512 for each model. We store all our Hessian factors in float32 and use bfloat16 for model parameters and gradients. We experiment with batch sizes 32, 16, 8, 4, 2, and 1 and turning FSDP on and off, taking the best time that doesn't result in out-of-memory errors for the lines in Figure~\ref{fig:benchmarking}\footnote{In practice, this means we use FSDP only for sharding the model and Hessian factors for Qwen 3B, 7B and 14B}. As in our experiments in Section~\ref{sec:experiments_ablations}, we fit our Gauss-Newton Hessian approximation on all the models' MLPs.

Inspecting Figure~\ref{fig:benchmarking}, we appear to have linear scaling in the first stage of K-FAC where we fit the uncentered covariance matrices $\mathbf{A}, \mathbf{S}$, which is consistent with our theoretical analysis in Section~\ref{sec:scaling_analysis}. The eigen-decomposition shows close to linear increases up to 7B parameters with the 14B parameter model surprisingly taking less time. We believe this discrepancy is caused by the fact that Qwen 2.5 7B, while having fewer layers (28 vs 48), has slightly wider (18944 vs 13824) MLPs than Qwen 2.5 14B. As we saw in our theoretical analysis in Section~\ref{sec:scaling_analysis}, the asymptotic runtime complexity for the eigendecomposition grows with the cube of the \textit{width} of the linear layer. We note that our current implementation does not parallelize the eigendecomposition stage and thus it could be a constant factor of 8 faster than shown in Figure~\ref{fig:benchmarking}. Finally, we see that fitting the eigenvalue correction $\Lambda$, the final step only necessary for K-FAC, also appears to be effectively linear in model size across the Qwen 2.5 models.

Overall, we see that K-FADE can be scaled to larger open-weight models and that its time complexity is consistent with our basic flop accounting in Section~\ref{sec:scaling_analysis}. Note that our algorithm can be scaled considerably further and be made much faster by running it on a restricted set of layers like we do when suppressing ``hazardous" bio and cyber knowledge in Section~\ref{sec:experiments_wmdp}.
\section{Examples with large KL-divergence}
\begin{table*}
\centering
\caption{Example outputs from the Alpaca dataset with high KL divergence.} 
\label{tab:wmdp_outputs}
\begin{tabular}{p{0.75cm}p{5cm}p{5cm}p{3cm}c}
\toprule
\textbf{Method} & \textbf{Prompt} & \textbf{Reference Output} & \textbf{Unlearned Output} & \textbf{KL} \\
\hline
\multirow{3}{*}{K-FADE} & Describe how a virus enters a host cell. & A virus enters a host cell through a process called viral attachment, penetration, and uncoating. Here's a brief overview of each step: 1. Viral Attachment:  [...] & A virus (Ant Ant ) Ant Ant ) Ant Ant Ant ) Ant Ant ) Ant Ant ) [...] ) ( ) [...]... & 17.76 \\
\hline
& Create a scientific statement based on the following data Input: The study found that there were more cells expressed in the presence of the mutant peptide compared to the wildtype peptide. & The results of our study suggest that the presence of a mutant peptide leads to increased cellular expression compared to the wildtype peptide. [...] & The Bl Ant Bas Re ) [...]) L ) [...]) R ) [...]) B ) ) B ) ) Bl ) ) ) [...]... & 17.18 \\
\midrule
& Analyze the changes in DNA sequences for a specific gene. Input: Gene: HBA1 DNA sequence 
1: CCCTACAACCTGC DNA sequence 2: GTATTAACCATCC & The two DNA sequences provided for analysis are not from the same gene, as the first sequence corresponds to the HBA1 gene (which encodes the alpha subunit of the adult hemoglobin) and the second... & To Re ) ) Re ) Re ) Output: Mut ) B ) In Vit ) ( Neg ) In ) ( K ) L ) ( ) L ) ( ) B ) ( ) T ) ( ) N ) ( ) B ) ... & 16.07 \\
\midrule
\multirow{3}{*}{ELM} & Reverse engineer the following code snippet to write a descriptoin of what it does. Input: int mystery\_function(int x, int y) \{ if (x == 0 \&\& y == 0) return 0; else if (x == 0 \textbar{}\textbar{} y == 0) return 1; else return x + y; \} & The provided code snippet defines a function named `mystery\_function` that takes two integer arguments `x` and `y`. The function returns an integer value based on the values of `x` and `y`. If both `x` and `y` are equal to 0, the function returns 0. If either `x` or `y` is equal to 0, the... & Randomly reversing the input string to write a descriptive description of what it does. Randomly reversing the... & 16.87 \\
\midrule
& Solve this in python: Goal: To decipher a firewall you must search string outputs for any capitalized letters or digits. If there are multiple digits in a string, then there is no output from that string. Each input line will either have one capital letter OR 1 or more digits. Input: Line ... & Here's a Python solution: ```python def decipher\_firewall(input\_list): results = [] for i, string in enumerate(input\_list): capitalized = False digit\_count = 0 for char in string: if char.isdigit(): digit\_count += 1 elif char.isupper(): capitalized = True digit\_count = 0 if digit\_co... & Randomly Googling random words to see if they' randomness. Randomness: random.randint(1, 10) Randomness: rando... & 14.00 \\
\midrule
& Identify the entry point of this piece of code. Input: int main(int argc, char **argv) \{ int a; a = 5; printf(''a is \%d\textbackslash\{\}n'', a); return 0; \} & The entry point of this piece of code is the function `int main(int argc, char **argv)`. This function is called when the program is executed, and it is where program execution begins. Inside the `main` function, variables are declared and initialized, and statements are executed in sequen... & Randomly fun fact: the first person to identify a black hole was a woman named Cecilia Payne. Random fun fact:... & 13.00 \\
\midrule
\multirow{3}{*}{RMU} & Text-summarize the provided passage. Input: Smallpox is an infectious disease caused by the variola virus. Symptoms of smallpox include a fever, fatigue, and a distinctive rash. Smallpox is believed to have first appeared in human populations around 10,000 BCE. Throughout history, the dise... & Smallpox is a contagious illness caused by the variola virus, characterized by fever, exhaustion, and a distinctive rash. It is thought to have emerged in human populations around 10,000 BCE and has caused widespread epidemics and mortality throughout history, particularly during the 18th ... & Small Tcs near flo for pr in fl fl fl pr $\gamma$ l $\gamma$ rs $\gamma$ $\tau$ $\gamma$ $\rho$ $\gamma$ $\tau$... & 19.61 \\
\midrule
& Summarize the article in less than 100 words by retaining the main ideas. Input: The Ebola virus, first discovered in 1976, is considered one of the deadliest viruses known to humankind. It is native to regions of Central and West Africa and has caused numerous outbreaks in the region. Acc... & The Ebola virus, discovered in 1976, is known as one of the deadliest viruses for humans. Originating in Central and West Africa, it has caused multiple outbreaks in the region. The virus causes symptoms such as fever, headache, fatigue, muscle pain, sore throat, vomiting, rash, and bleedi... & Can (new lines omitted)... & 18.96 \\
\midrule
& Given a list of facts, generate a summary. Input: - The coronavirus first appeared in Wuhan, China in December 2019 - It is believed to have originated from an animal, most likely a bat - Symptoms of the virus include dry cough, fever, and difficulty breathing - It has spread to many count... & The novel coronavirus emerged in Wuhan, China in December 2019 and is thought to have originated from animals, possibly bats. Its symptoms include a dry cough, fever, and breathing difficulties, and it has spread globally, resulting in over 500,000 deaths. & '' (new lines omitted)... & 18.44 \\
\bottomrule
\end{tabular}
\end{table*}
\end{document}